\documentclass[11pt]{article}

\usepackage[round]{natbib}
\usepackage[utf8]{inputenc} 
\usepackage[T1]{fontenc}    
\usepackage{url}            
\usepackage{booktabs}       
\usepackage{amsfonts}       
\usepackage{nicefrac}       
\usepackage{microtype}      
\usepackage{xcolor}         
\usepackage{hyperref}       
\usepackage{amsthm}

\usepackage[top=1in, left=1in, right=1in, bottom=1in]{geometry}

\usepackage{amsfonts}       
\usepackage{subcaption}        
\usepackage{graphicx}
\usepackage[ruled]{algorithm2e}
\usepackage{algpseudocode}
\usepackage{multirow}
\usepackage{amssymb}
\usepackage{pifont}
\usepackage{wrapfig,lipsum}
\usepackage{enumitem}

\usepackage{bbm}
\usepackage{authblk}
\usepackage{makecell}
\usepackage{multicol}

\theoremstyle{plain}
\newtheorem{theorem}{Theorem}[section]

\theoremstyle{definition}

\newcommand{\egnn}[1]{\texttt{EGNN} #1}

\usepackage{amsmath,amsfonts,bm}

\newcommand{\relu}[1]{\textrm{ReLU} #1}

\def\eqref#1{equation~(\ref{#1})}

\def\1{\bm{1}}

\def\vh{{\bm{h}}}

\def\vx{{\bm{x}}}
\def\vy{{\bm{y}}}

\def\mA{{\bm{A}}}

\def\mD{{\bm{D}}}

\def\mH{{\bm{H}}}
\def\mI{{\bm{I}}}

\def\mX{{\bm{X}}}

\def\mPhi{{\bm{\Phi}}}

\def\mTheta{{\bm{\Theta}}}

\DeclareMathAlphabet{\mathsfit}{\encodingdefault}{\sfdefault}{m}{sl}
\SetMathAlphabet{\mathsfit}{bold}{\encodingdefault}{\sfdefault}{bx}{n}

\def\gE{{\mathcal{E}}}

\def\gG{{\mathcal{G}}}

\def\gV{{\mathcal{V}}}

\newcommand{\R}{\mathbb{R}}

\DeclareMathOperator*{\argmax}{arg\,max}

\makeatletter
\newcommand{\printfnsymbol}[1]{%
  \textsuperscript{\@fnsymbol{#1}}%
}
\makeatother

\title{Editable Graph Neural Network for Node Classifications}

\author[1]{Zirui Liu\thanks{Equal contribution. The order of authors is determined by flipping a coin.}}
\author[2]{Zhimeng Jiang\printfnsymbol{1}}
\author[1]{Shaochen Zhong}
\author[1]{Kaixiong Zhou}
\author[3]{Li Li}
\author[3]{Rui Chen}
\author[3]{Soo-Hyun Choi}
\author[1]{Xia Hu}
\affil[1]{Department of Computer Science, Rice University}
\affil[2]{Department of Computer Science, Texas A\&M University}
\affil[3]{Samsung Electronics America}
\affil[ ]{{\texttt{\{Zirui.Liu, Shaochen.Zhong, Kaixiong.Zhou, Xia.Hu\}@rice.edu}, \texttt{zhimengj@tamu.edu},
\affil[ ]\texttt{\{li.li1, rui.chen1,sh9.choi\}@samsung.com}}}

\date{}

\begin{document}

\maketitle

\begin{abstract}
Despite Graph Neural Networks (GNNs) have achieved prominent success in many graph-based learning problem, such as credit risk assessment in financial networks and fake news detection in social networks.
However, the trained GNNs still make errors and these errors may cause serious negative impact on society.
\textit{Model editing}, which corrects the model behavior on wrongly predicted target samples while leaving model predictions unchanged on unrelated samples, has garnered significant interest in the fields of computer vision and natural language processing. However, model editing for graph neural networks (GNNs) is rarely explored, despite GNNs' widespread applicability. 
To fill the gap, we first observe that existing model editing methods significantly deteriorate prediction accuracy (up to $50\%$ accuracy drop) in GNNs while a slight accuracy drop in multi-layer perception (MLP). 
The rationale behind this observation is that the node aggregation in GNNs will spread the editing effect throughout the whole graph.
This propagation pushes the node representation far from its original one.
Motivated by this observation, we propose \underline{E}ditable \underline{G}raph \underline{N}eural \underline{N}etworks (EGNN), a neighbor propagation-free approach to correct the model prediction on misclassified nodes. 
Specifically, EGNN simply stitches an MLP to the underlying GNNs, where the weights of GNNs are frozen during model editing. In this way, EGNN disables the propagation during editing while still utilizing the neighbor propagation scheme for node prediction to obtain satisfactory results. 
Experiments demonstrate that EGNN outperforms existing baselines in terms of effectiveness (correcting wrong predictions with lower accuracy drop), generalizability (correcting wrong predictions for other similar nodes), and efficiency (low training time and memory) on various graph datasets.

\end{abstract}

\section{Introduction}




Graph Neural Networks (GNNs) have achieved prominent results in learning features and topology of graph data \citep{ying2018graph, gsage, ling2023learning, gsaint, ogb, zhoutable2graph, fmp, han2022geometric, gmixup,ling2023graph, DBLP:conf/nips/DuanLWZZCHW22, DBLP:journals/corr/abs-2108-13555}. 
Based on spatial message passing, GNNs learn each node through aggregating  representations of its neighbors and the node itself recursively. 
Once trained, the model is typically deployed as static artifacts to make decisions on a wide range of tasks, such as credit risk assessment in financial networks \citep{petrone2018dynamic} and fake news detection in social networks \citep{shu2017fake}. 
However, the cost of making a wrong decision could be higher in these graph applications. Over-trusted creditworthiness on borrowers can lead to severe loss for lenders, and failure detection of fake news has a serious negative impact on society. 

Ideally, it is desirable to correct these serious errors and generalize corrections to similar mistakes, while preserving the model's prediction accuracy on unrelated input samples.
To obtain generalization ability for similar samples, the most prevalent method is to fine-tune the model with a new label on the single example to be corrected.
However, this approach often spoils the model prediction on other unrelated samples.
To cope with the challenge, many model editing frameworks have been proposed to adjust model behaviors by correcting errors as they appear \citep{enn, mitchell2021fast, mitchell2022memory, de2021editing}.
Specifically, these editors usually require an additional training phase to help the model ``prepare'' for the editing process before applying any edits \citep{enn, mitchell2021fast, mitchell2022memory, de2021editing}.

Although model editing has shown promise to modify vision and language models, to the best of our knowledge, there is no existing work tackling the critical mistakes in graph data. 
Despite the straightforward concept, it is challenging to efficiently change GNNs' behaviors on the massively connected nodes. 
First, due to the message-passing mechanism in GNNs, editing the model behavior on a single node can propagate changes across the entire graph, significantly altering the node's original representation, which may destroy the prediction performance on the training dataset. 
Therefore, compared to the neural networks for computer vision or natural language processing, it is harder to maintain the model prediction on other input samples.
Second, unlike other types of neural networks, the input nodes are connected in the graph domain. 
Thus, when editing the model prediction on a single node using gradient descent, the representation of each node in the whole graph is required \citep{exact, han2023mlpinit, gsage}. 
This distinction introduces complexity and computational challenges when making targeted adjustments to GNNs, especially on large graphs.


In this work, we delve into studying the graph model editing problem, which is more challenging than the independent sample edits. 
We first observe the existing editors significantly harm the overall node classification accuracy although the misclassified nodes are corrected. 
The test accuracy drop is up to $50\%$, which prevents GNNs from being practically deployed. 
We experimentally study the rationale behind this observation from the lens of loss landscapes.
Specifically, we visualize the loss landscape of the Kullback-Leibler (KL) divergence between node embeddings obtained before and after the model editing process in GNNs.
We found that a slight weight perturbation can significantly enlarge the KL divergence.
In contrast, other types of neural networks, such as Multi-Layer Perceptrons (MLPs), exhibit a much flatter region of the KL loss landscape and display greater robustness against weight variations.
Such observations align with our viewpoint that after editing on misclassified samples, GNNs are prone to widely propagating the editing effect and affecting the remaining nodes. 

Based on the sharp loss landscape of model editing in GNNs, we propose Editable Graph Neural Network (\egnn), a neighbor propagation-free approach to correct the model prediction on the graph data. 
Specifically, suppose we have a well-trained GNN and we found want to correct its prediction on some of the misclassified nodes.
\egnn stitches a randomly initialized MLP to the trained GNN.
We then train the MLP for a few iterations to ensure that it does not significantly alter the model's prediction.
When performing the edit, 
we only update the parameter of the stitched MLP while freezing the parameter of GNNs during the model editing process. 
In particular, the node embeddings from GNNs are first inferred offline.
Then MLP learns an additional representation, which is then combined with the fixed embeddings inferred from GNNs to make the final prediction.
When a misclassified node is received, the gradient is back propagated to update the parameters of MLP instead of GNNs'. 
In this way, we decouple the \textit{neighbor propagation process} of learning the structure-aware node embeddings from the \textit{model editing process} of correcting the misclassified nodes. 
Thus, \egnn disables the propagation during editing while still utilizing the neighbor propagation scheme for node prediction to obtain satisfactory results.
Compared to directly applying the existing model editing methods to GNNs:
\begin{itemize}[leftmargin=0.2cm, itemindent=.0cm, itemsep=0.0cm, topsep=0.0cm]
    \item We can leverage the GNNs' structure learning meanwhile avoiding the spreading edition errors to guarantee the overall node classification task. 
    \item The experimental results validate our solution which could address all the erroneous samples and 
    deliver up to \textbf{90\% improvement in overall accuracy}. 
    \item Via freezing GNNs' part, \egnn is scalable to address misclassified nodes in the million-size graphs. We save more than $2\times$ in terms of memory footprint and model editing time. 
\end{itemize}

\section{Preliminary}
\paragraph{Graph Neural Networks.}
Let $\gG=(\gV,\gE)$ be an undirected graph with $\gV=(v_1,\cdots,v_{|\gV|})$ and $\gE=(e_1,\cdots,e_{|\gE|})$ being the set of nodes and edges, respectively.
Let $\mX\in\R^{|\gV|\times d}$ be the node feature matrix.
$\mA \in \R^{|\gV|\times|\gV|}$ is the graph adjacency matrix, where $\mA_{i,j}=1$ if $(v_i, v_j)\in \gE$ else $\mA_{i,j}=0$.
$\Tilde{\mA} = \Tilde{\mD}^{-\frac{1}{2}}(\mA+\mI)\Tilde{\mD}^{-\frac{1}{2}}$ is the normalized adjacency matrix, where $\Tilde{\mD}$ is the degree matrix of $\mA+\mI$.
In this work, we are mostly interested in the task of node
classification, where each node $v\in\gV$ is associated with a
label $y_v$, and the goal is to learn a representation $\vh_v$ from which $y_v$ can be easily predicted. 
To obtain such a representation, GNNs follow a neural message passing scheme \citep{gcn}.
Specifically, GNNs recursively update the representation of a node by aggregating representations of its neighbors.
For example, the $l^{\mathrm{th}}$ Graph Convolutional Network (GCN) layer~\citep{gcn} can be defined as:
\begin{equation}
\label{eq: gcnconv}
    \mH^{(l+1)}=\relu(\Tilde{\mA}\mH^{(l)}\mTheta^{(l)}),
\end{equation}
where $\mH^{(l)}$ is the node embedding matrix containing the $\vh_v$ for each node $v$ at the $l^{\mathrm{th}}$ layer and $\mH^{(0)}=\mX$. 
$\mTheta^{(l)}$ is the weight matrix of the $l^{\mathrm{th}}$ layer.

\paragraph{The Model Editing Problem.}
The goal of model editing is to alter a base model’s output for a misclassified sample $x_e$ as well as its similar samples via model finetuning only using a single pair of input $x_e$ and desired output $y_e$ while leaving model behavior on unrelated inputs intact \citep{enn, mitchell2021fast, mitchell2022memory}. We are the first to propose the model editing problem in graph data, where the decision faults on a small number of critical nodes can lead to significant financial loss and/or fairness concerns.  For the node classification, suppose a well-trained GNN incorrectly predicts a specific node.
\textbf{Model editing} is used to correct the undesirable prediction behavior for that node by using the node's features and desired label to update the model. Ideally, the model editing ensures that the updated model makes accurate predictions for the specific node and its similar samples while maintaining the model's original behavior for the remaining unrelated inputs. Some model editors, such as the one presented in this paper, require a training phase before they can be used for editing.  




\section{Proposed Methods}

\label{sec: motivation}
In this section, we first empirically show vanilla model editing performs extremely worse for GNNs compared with MLPs due to node propagation (Section \ref{sec: cryGNN}). 
Intuitively, due to the message-passing mechanism in GNNs, editing the model behavior on a single node can propagate changes across the entire graph, significantly altering the node's original representation.
Then through visualizing the loss landscape, we found that for GNNs, even a slight weight perturbation, the node representation will be far away from the original one (Section \ref{sec: losslandscapeGNN}).
Based on the observation, we propose a propagation-free GNN editing method called \egnn (Section \ref{sec: solution}).


\subsection{Motivation: Model Editing may Cry in GNNs}\label{sec: cryGNN}
\noindent

\begin{table}
\centering
\captionsetup{skip=5pt} 
\vspace{-1em}
\caption{The test accuracy ($\%$) before (``w./o. edit'') and after editing (``w./ edit'') on one single data point. $\Delta$ Acc is the accuracy drop before and after performing the edit.
All results are averaged over 50 simultaneous model edits. 
The best result is highlighted by \bf{bold faces}.}
\label{tab: prelim_exp_res}
\begin{tabular}{ccccc} 

\hline
                            &               & GCN               & GraphSAGE         & MLP                          \\ 
\hline
\multirow{3}{*}{Cora}       & ~w./o. edit   & \textbf{89.4}     & 86.6              & 71.8                         \\
                            & w./ edit      & \textbf{84.36}    & 82.06             & 68.33                        \\
                            & $\Delta$ Acc.     & 5.03$\downarrow$  & 4.53$\downarrow$  & \textbf{3.46} $\downarrow$   \\ 
\hline
\multirow{3}{*}{Flickr}     & ~w./o. edit   & \textbf{51.19}    & 49.03             & 46.77                        \\
                            & w./ edit      & 13.94             & 17.15             & \textbf{36.68}               \\
                            & $\Delta$ Acc. & 37.25$\downarrow$ & 31.88$\downarrow$ & \textbf{10.08} $\downarrow$  \\ 
\hline
\multirow{3}{*}{Reddit}     & ~w./o. edit   & 95.52             & \textbf{96.55}    & 72.41                        \\
                            & w./ edit      & \textbf{75.20}    & 55.85             & 69.86                        \\
                            & $\Delta$ Acc. & 20.32$\downarrow$ & 40.70$\downarrow$ & \textbf{2.54} $\downarrow$   \\ 
\hline
\multirow{3}{*}{ogbn-arxiv} & w./o. edit    & \textbf{70.20}    & 68.38             & 52.65                        \\
                            & w./ edit      & 23.70             & 19.06             & \textbf{45.15}               \\
                            & $\Delta$ Acc. & 46.49$\downarrow$ & 49.31$\downarrow$ & \textbf{7.52}$\downarrow$    \\
\hline
\end{tabular}
\end{table}

\textbf{Setting:}
We train GCN, GraphSAGE, and MLP on Cora, Flickr, Reddit, and ogbn-arxiv, respectively, following the training setup as described in Section \ref{sec: exp}.
To evaluate the difficulty of editing, \emph{we ensured that the node to be edited was not present during training}, meaning that \emph{the models were trained inductively}.
Specifically, we trained the model on a subgraph containing only the training node and evaluated its performance on the validation and test set of nodes.
Next, we selected a misclassified node from the validation set and applied gradient descent only on that node until the model made a correct prediction for it.
Following previous work \citep{enn, mitchell2022memory}, we perform 50 independent edits and report the averaged test accuracy before and after performing a single edit.


\noindent
\textbf{Results:} As shown in Table \ref{tab: prelim_exp_res}, we observe that \textbf{(1)} GNNs consistently outperform MLP on all the graph datasets before editing. This is consistent with the previous graph analysis results, where the neural message passing involved in GNNs extracts the graph topology to benefit the node representation learning and thereby the classification accuracy. 
\textbf{(2)} After editing, the accuracy drop of GNNs is significantly larger than that of MLP. 
For example, GraphSAGE has an almost 50\% drop in test accuracy on ogbn-arxiv after editing even a single point. MLP with editing even delivers higher overall accuracies on Flickr and ogbn-arxiv compared with GNN-based approaches. One of the intuitive explanations is the slightly fine-tuned weights in MLP mainly affect the target node, instead of other unrelated samples. However, due to the message-passing mechanism in GNNs, the edited node representation can be propagated over the whole graph and thus change the decisions on a large area of nodes. These comparison results reveal the unique challenge in editing the correlated nodes with GNNs, compared with the conventional neural networks working on isolated samples. 
\textbf{(3)} After editing, the test accuracy of GCN, GraphSAGE, and MLP become too low to be practically deployed. This is quite different to the model editing problems in computer vision and natural language processing, where the modified models only suffer an acceptable accuracy drop. 

\subsection{Sharp Locality of GNNs through Loss Landscape} \label{sec: losslandscapeGNN}
Intuitively, due to the message-passing mechanism in GNNs, editing the model behavior for a single node can cause the editing effect to propagate across the entire graph. This propagation pushes the node representation far from its original one.
\textbf{Thus, we hypothesized that the difficulty in editing GNNs as being due to the neighbor propagation of GNNs.}
The model editing aims to correct the prediction of the misclassified node using the cross-entropy loss of desired label. 
Intuitively, the large accuracy drop can be interpreted as the low model prediction similarity before and after model editing, named as the locality.

\begin{figure}[t!]
    \centering
        \vspace{-1em}
    \includegraphics[width=0.99\linewidth]{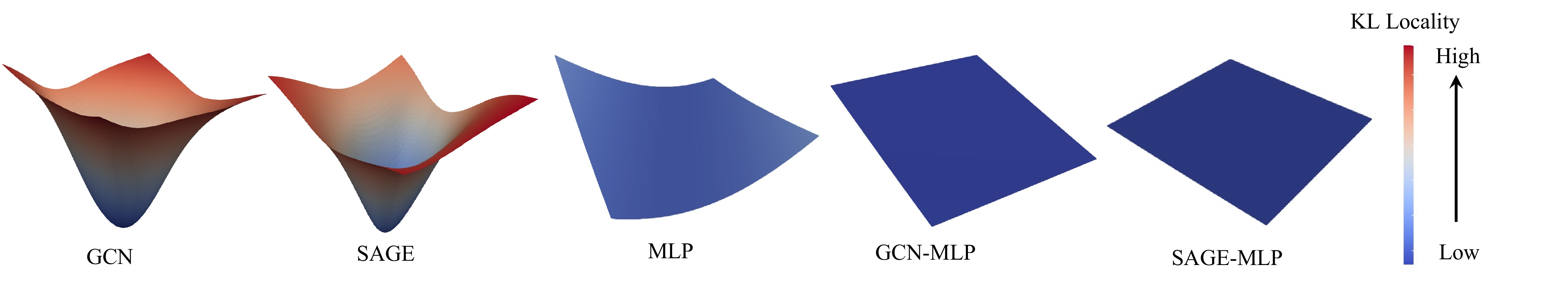}
    \vspace{-.5em}
    \caption{The loss landscape of various model architectures on Cora dataset.
    Similar results can be found in Appendix \ref{app: more_exp_res}}
    \vspace{-1em}
    \label{fig: landscape_1}
\end{figure}

To quantitatively measure the locality, we use the metric of KL divergence between the node representations learned before and after model editing. The higher KL divergence means after editing, the node representation is far away from the original one. 
In other words, the higher KL divergence implies poor model locality, which is undesirable in the context of model editing. Particularly, we visualize the locality loss landscape for Cora dataset in Figure~\ref{fig: landscape_1}. 
We observe several \textbf{insights:} (1) GNNs (e.g., GCN and GraphSAGE) suffer from a much sharper loss landscape. Even slightly editing the weights, KL divergence loss is dramatically enhanced. 
That means GNNs are hard to be fine-tuned while keeping the locality.
(2) MLP shows a flatter loss landscape and demonstrates much better locality to preserve overall node representations. 
This is consistent to the accuracy analysis in Table~\ref{tab: prelim_exp_res}, where the accuracy drop of MLP is smaller. 

To deeply understand why model editing fails to work in GNNs, we also provide a pilot theoretical analysis on the KL locality difference between before/after model editing for one-layer GCN and MLP in Appendix \ref{app: analsis}. 
We theoretically show that when model editing corrects the model predictions on misclassified nodes, GNNs are susceptible to altering the predictions on other connected nodes. This phenomenon results in an increased KL divergence difference.

\subsection{\texorpdfstring{$\egnn$}: Neighbor Propagation Free GNN Editing}\label{sec: solution}

\begin{figure}[ht!]
    \centering
    \includegraphics[width=1.0\linewidth]{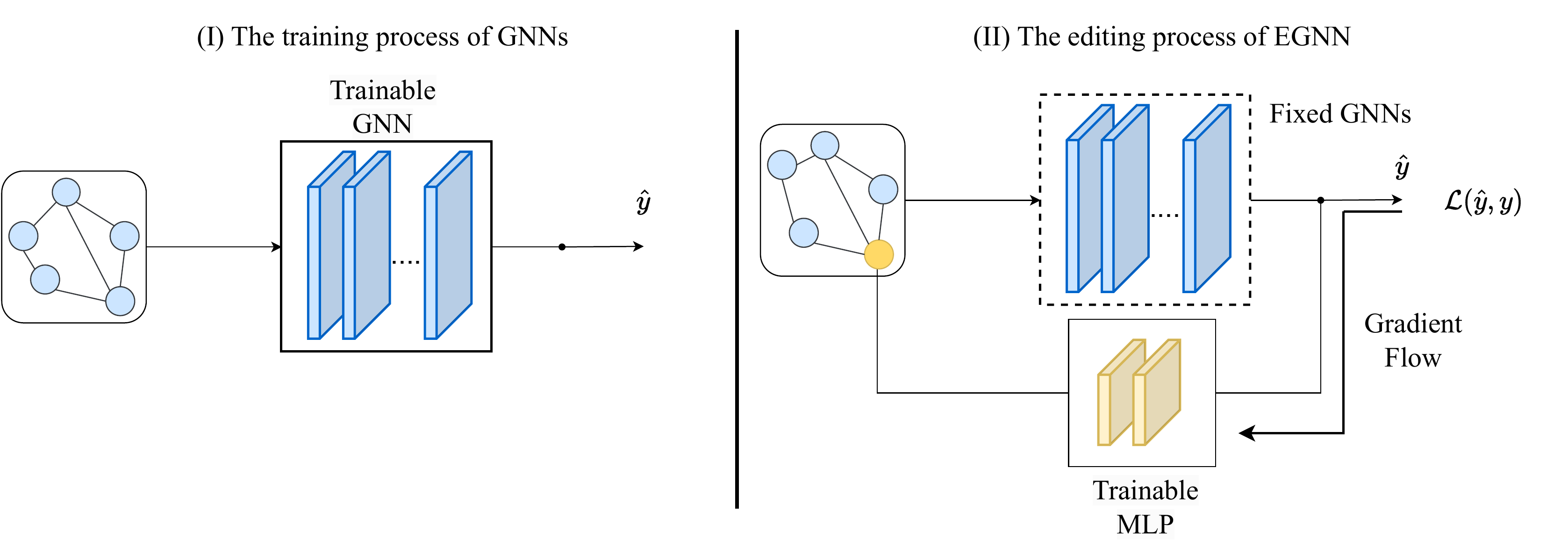}
    \vspace{-1em}
    \caption{The overview of \egnn.
    We fix the backbone of GNNs (in blue), while only update the small MLPs (in orange) during editing.
    The wrongly predicted nodes are highlighted with orange color, and the MLP only require its node feature.}
        \vspace{-.5em}
    \label{fig: framework}
\end{figure}

In our previous analysis, we hypothesized that the difficulty in editing GNNs as being due to the neighbor propagation. 
However, as Table \ref{tab: prelim_exp_res} suggested, the neighbor propagation is necessary for obtaining good performance on graph datasets. On the other hand, MLP could stabilize most of the node representations during model editing although it has worse node classification capability. Thus, we need to find a way to ``disable'' the propagation during editing while still utilizing the neighbor propagation scheme for node prediction to obtain satisfactory results. 
Following the motivation, we propose to combine a compact MLP to the well-trained GNN and only modify the MLP during editing. 
In this way, 
we can correct the model's predictions through this additional MLP while freezing the neighbor propagation.
Meanwhile during inference, both the GNN and MLP are used together for prediction in tandem to harness the full potential of GNNs for prediction.
The whole algorithm is shown in Algorithm \ref{algo: whole}.

\begin{algorithm}[H]  
  \SetKwProg{myProcedure}{procedure}{\string:}{end procedure}
  \myProcedure{\textsc{MLP training procedure}}{
\KwIn{MLP $g_\mPhi$, dataset $\mathcal{D}$, the node embedding $\vh_v$ for each node $v$ in $\mathcal{D}$}
\For{$t=1,\cdots,T$}{
    Sample $\vx_v$, $y_v\sim \mathcal{D}^{\text{train}}$\\
    $\mathcal{L}_{\text{loc}}=\text{KL}(\vh_v +g_\mPhi(\vx_v)|| \vh_v)$\\
    $\mathcal{L}_{\text{task}}=-\log p_{\mPhi}(y_v|\vh_v+g_\mPhi(\vx_v))$\\
    $\mathcal{L}=\mathcal{L}_{\text{task}}$ + $\alpha\mathcal{L}_{\text{loc}}$\\
    $\mPhi\leftarrow Adam(\mPhi, \nabla\mathcal{L})$
}
  }
  \myProcedure{\textsc{$\egnn$ edit procedure}}{
  \KwIn{data pair $x_e, y_e$ to be edited, the node embedding $\vh_e$ for node $e$}
$\hat{y}=\argmax_y p_{\mPhi}(y|\vx_e, \vh_e)$\\
\While{$\hat{y}\neq y_v$}{
    $\mathcal{L}=-\log p_{\mPhi}(y|\vx_e, \vh_e)$\\
    $\mPhi\leftarrow Adam(\mPhi, \nabla\mathcal{L})$
}
}
  \caption{Proposed $\egnn$}
  \label{algo: whole}
\end{algorithm}

\textbf{Before editing.} We first stitch a randomly initialized compact MLP to the trained GNN.
To mitigate the potential impact of random initialization on the model's prediction, we introduce a training procedure for the stitched MLP, as outlined in Algorithm \ref{algo: whole} ``\textsc{MLP training procedure}'': we train the MLP for a few iterations to ensure that it does not significantly alter the model's prediction.
By freezing GNN's weights, we first get the node embedding $\vh_v$ at the last layer of the trained GNN by running a single forward pass.
We then stitch the MLP with the trained GNNs. Mathematically, we denote the MLP as $g_\mPhi$ where  $\mPhi$ is the parameters of MLP.
For a given input sample $\vx_v, \vy_v$, the model output now becomes $\vh_v+g_\mPhi(\vx_v)$.
We calculate two loss based on the prediction, i.e., the task-specific loss $\mathcal{L}_{\text{task}}$ and the locality loss $\mathcal{L}_{\text{loc}}$.
Namely,
\begin{align}
    \mathcal{L}_{\text{task}} &=-\log p_{\mPhi}(y_v|\vh_v+g_\mPhi(\vx_v)), \nonumber \\
    \mathcal{L}_{\text{loc}} &= \text{KL}(\vh_v +g_\mPhi(\vx_v)|| \vh_v), \nonumber
\end{align}

where $\vh_v+g_\mPhi(\vx_v)$ is the model prediction with the additional MLP and $p_{\mPhi}(y_v|\vh_v+g_\mPhi(\vx_v))$ is the probability of class $y_v$ given by the model. 
$\mathcal{L}_{\text{task}}$ is the cross-entropy between the model prediction and label.
$\mathcal{L}_{\text{loc}}$ is the locality loss, which equals KL divergence between the original prediction $\vh_v$ and the prediction with the additional MLP $\vh_v+g_\mPhi(\vx_v)$.
The final loss $\mathcal{L}$ is the weighted combination of two parts, i.e., $\mathcal{L}=\mathcal{L}_{\text{task}}$ + $\alpha\mathcal{L}_{\text{loc}}$ where $\alpha$ is the weight for the locality loss.
$\mathcal{L}$ is used to guide the MLP to fit the task while keep the model prediction unchanged.

\textbf{When editing.} \textbf{\egnn freezes the model parameters of GNN and only updates the parameters of MLP.}
Specifically, as outlined in Algorithm \ref{algo: whole} ``\textsc{\egnn Edit Procedure}'', we update the parameters of MLP until the model prediction for the misclassified sample is corrected.
Since MLP only relies on the node features, we can easily perform these updates in mini-batches, which enables us to edit GNNs on large graphs. 

Lastly, we visualize the KL locality loss landscape of EGNN (including GCN-MLP and SAGE-MLP) in Figure~\ref{fig: landscape_1}. It is seen that the proposed EGNN shows the most flattened loss landscape than MLP and GNNs, which implied that EGNN can preserve overall node representations better than other model architectures.

\section{Related Work and Discussion}
Due to the page limit, below we discuss the related work on model editing.
We also discuss the limitation in Appendix \ref{app:limit}.
\paragraph{Model Editing.} Many approaches have been proposed for model editing. The most straightforward method adopts standard fine-tuning to update model parameters based on misclassified samples while preserving model locality via constraining parameters travel distance in model weight space~\citep{zhu2020modifying, sotoudeh2019correcting}. Work~\citep{sinitsin2020editable} introduces meta-learning to find a pre-trained model with rapid and easy finetuned ability for model editing. Another way to facilitate model editing relies on external learned editors to modify model editing considering several constraints~\citep{mitchell2021fast,hase2021language,de2021editing,mitchell2022memory}.  The editing of the activation map is proposed to correct misclassified samples in~\citep{dai2021knowledge,meng2022locating} due to the belief of knowledge attributed to model neurons. While all these works either update base model parameters or import external separate modules for model prediction to induce desired prediction change, 
the considered data is i.i.d. and may not work well in graph data due to essential node interaction during neighborhood propagation. In this paper, we propose EGNN, using a stitched MLP module to edit the output space of the base GNN model, for node classification tasks. The key insight behind this solution is the sharp locality of GNNs, i.e., the prediction of GNNs can be easily altered after model editing.  



\section{Experiments}
\label{sec: exp}

The experiments are designed to answer the following research questions.
\textbf{RQ1:} Can \egnn correct the wrong model prediction?
Moreover, what is the difference in accuracy before and after editing using \egnn?
\textbf{RQ2:} Can the edits generalize to correct the model prediction on other similar inputs? 
\textbf{RQ3:} What is the time and memory requirement of \egnn to perform the edits?

\subsection{Experimental Setup}
\paragraph{Datasets and Models.}
To evaluate \egnn, we adopt four small-scale and four large-scale graph benchmarks from different domains.
For small-scale datasets, we adopt Cora, A-computers \citep{shchur2018pitfalls}, A-photo \citep{shchur2018pitfalls}, and Coauthor-CS \citep{shchur2018pitfalls}.
For large-scale datasets, we adopt Reddit \citep{gsage}, Flickr \citep{gsaint}, \textit{ogbn-arxiv} \citep{ogb}, and \textit{ogbn-products} \citep{ogb}.
We integrate \egnn with two popular models: 
GCN \citep{gcn} and GraphSAGE \citep{gsage}.
\emph{To avoid creating confusion, 
GCN and GraphSAGE are all trained with the whole graph at each step}.
We evaluate \egnn under the \textbf{inductive setting}.
Namely, we trained the model on a subgraph containing only the training node and evaluated it on the whole graph.
Details about the hyperparameters and datasets are in Appendix \ref{app: exp_setting}.

\paragraph{Compared Methods.}
We compare our \egnn editor with the following two baselines: the vanilla gradient descent editor (GD) and Editable Neural Network editor (ENN) \citep{enn}.
GD is the same editor we used in our preliminary analysis in Section \ref{sec: motivation}. 
\textbf{We note that for other model editing, e.g., MEND \citep{mitchell2021fast}, SERAC \citep{mitchell2022memory} are tailored for NLP applications, which cannot be directly applied to the graph area}.
Specifically, GD applies the gradient descent \textbf{on the parameters of GNN} until the GNN makes right prediction.
ENN trains \textbf{the parameters of GNN} for a few steps to make it prepare for the following edits.
Then similar to GD editor, it applies the gradient descent \textbf{on the parameters of GNN} until the GNN makes right prediction.
For \egnn, we only train \textbf{the stitched MLP} for a few steps.
Then we only update \textbf{weights of MLP} during edits.
Detailed hyperparameters are listed in Appendix \ref{app: exp_setting}. 

\paragraph{Evaluation Metrics.}
Following previous work \citep{enn, mitchell2022memory, mitchell2021fast}, we evaluate the effectiveness of different methods by the following three metrics.
\textbf{DrawDown (DD)}, which is the mean absolute difference of test accuracy before and after performing an edit. 
A smaller drawdown indicates a better editor locality.
\textbf{Success Rate (SR)}, which is defined as the rate of edits, where the editor successfully corrects the model prediction. 
\textbf{Edit Time}, which is defined as the wall-clock time of a single edit that corrects the model prediction.


\subsection{The Effectiveness of \texorpdfstring{$\egnn$} in Editing GNNs}


\begin{table}[ht!]
        \centering
        \captionsetup{skip=5pt} 
        \caption{    The results on four small scale datasets after applying one single edit. 
        The reported number is averaged over 50 independent edits.    \textbf{SR} is the edit success rate,     \textbf{Acc} is the test accuracy after editing,     and \textbf{DD} are the test drawdown, respectively.    ``OOM'' is the out-of-memory error.}
        \label{tab: small_exp_res}
        \resizebox{\textwidth}{!}{
        \begin{tabular}{cccccccccccccc} 
        \hline
        \multirow{2}{*}{}                                                     & \multirow{2}{*}{Editor} & \multicolumn{3}{c}{Cora}                                                            & \multicolumn{3}{c}{A-computers}                                                      & \multicolumn{3}{c}{A-photo}                                                           & \multicolumn{3}{c}{Coauthor-CS}                                                        \\ 
        \cline{3-14}
                                                                              &                         & Acc$\uparrow$       & DD$\downarrow$                                 & SR$\uparrow$ & Acc$\uparrow$       & DD$\downarrow$                                  & SR$\uparrow$ & Acc$\uparrow$        & DD$\downarrow$                                  & SR$\uparrow$ & Acc$\uparrow$       & DD$\downarrow$                                 & SR$\uparrow$  \\ 
        \hline
        \multirow{3}{*}{GCN}                                                  & GD                      & 84.37±5.84          & 5.03±6.40                                      & 1.0          & 44.78±22.41         & 43.09±22.32                                     & 1.0          & 28.70±21.26          & 65.08±20.13                                     & 1.0          & 91.07±3.23          & 3.30±2.22                                      & 1.0           \\
                                                                              & ENN                     & 37.16±3.80          & \textcolor[rgb]{0.122,0.125,0.141}{52.24}±4.76 & 1.0          & 15.51±10.99         & \textcolor[rgb]{0.122,0.125,0.141}{72.36}±10.87 & 1.0          & 16.71±14.81          & \textcolor[rgb]{0.122,0.125,0.141}{77.07}±15.20 & 1.0          & 4.94±3.78           & \textcolor[rgb]{0.122,0.125,0.141}{89.43}±3.34 & 1.0           \\
                                                                              &  \egnn                       & \textbf{87.80}±2.34 & \textbf{1.80}±2.13                             & \textbf{1.0} & \textbf{82.85}±5.20 & \textbf{2.32}±5.11                              & 0.98         & \textbf{91.97}±5.85  & \textbf{2.39}±5.34                              & \textbf{1.0} & \textbf{94.54}±0.07 & ~-\textbf{0.17}±0.07                                    & \textbf{1.0}  \\ 
        \hline
        \multirow{3}{*}{\begin{tabular}[c]{@{}c@{}}Graph-\\SAGE\end{tabular}} & GD                      & 82.06±4.33          & 4.54±5.32                                      & 1.0          & 21.68±20.98         & 61.15±20.33                                     & 1.0          & 38.98±30.24          & 55.32±29.35                                     & 1.0          & 90.15±5.58          & 5.01±5.32                                      & 1.0           \\
                                                                              & ENN                     & 33.16±1.45          & \textcolor[rgb]{0.122,0.125,0.141}{53.44}±2.23 & 1.0          & 16.89±16.98         & 65.94±16.75                                     & 1.0          & 15.06±11.92          & \textcolor[rgb]{0.122,0.125,0.141}{79.24}±11.25 & 1.0          & 13.71±2.73          & \textcolor[rgb]{0.122,0.125,0.141}{81.45}±2.11 & 1.0           \\
                                                                              & \egnn                        & \textbf{85.65}±2.23 & \textbf{0.55}±1.26                             & \textbf{1.0} & \textbf{84.34}±4.84 & \textbf{2.72}±5.03                              & 0.94         & \textbf{~92.53}±2.90 & \textbf{1.83}±3.22                              & \textbf{1.0} & \textbf{95.27}±0.08 & ~-\textbf{0.01}±0.10                                    & \textbf{1.0}  \\
        \hline
        \end{tabular}}
        \end{table}
       
\begin{table}
        \centering
        \caption{    The results on four large scale datasets after applying one single edit. ``OOM'' is the out-of-memory error.}
        \label{tab: main_exp_res}
            \resizebox{\textwidth}{!}{
        \begin{tabular}{cccccccccccccc} 
        \hline
        \multirow{2}{*}{}                                                     & \multirow{2}{*}{Editor} & \multicolumn{3}{c}{Flickr}                              & \multicolumn{3}{c}{Reddit}                             & \multicolumn{3}{c}{\begin{tabular}[c]{@{}c@{}}ogbn-\\arxiv\end{tabular}} & \multicolumn{3}{c}{\begin{tabular}[c]{@{}c@{}}ogbn-\\products\end{tabular}}  \\ 
        \cline{3-14}
                                                                              &                         & Acc$\uparrow$       & DD$\downarrow$     & SR$\uparrow$ & Acc$\uparrow$      & DD$\downarrow$     & SR$\uparrow$ & Acc$\uparrow$      & DD$\downarrow$    & SR$\uparrow$                    & Acc$\uparrow$       & DD$\downarrow$ & SR$\uparrow$                          \\ 
        \hline
        \multirow{3}{*}{GCN}                                                  & GD                      & 13.95±11.0          & 37.25±10.2         & 1.0          & 75.20±12.3         & 20.32±11.3         & 1.0          & 23.71±16.9         & 46.50±14.9        & 1.0                             & OOM                 & OOM            & 0                                     \\
                                                                              & ENN                     & 25.82±14.9          & 25.38±16.9         & 1.0          & 11.16±5.1          & 84.36±3.1          & 1.0          & 16.59±7.7          & 53.62±6.7         & 1.0                             & OOM                 & OOM            & 0                                     \\
                                                                              & \egnn                   & \textbf{44.91}±12.2 & \textbf{6.34}±10.3 & \textbf{1.0} & \textbf{94.46}±0.4 & \textbf{1.03}±0.6  & \textbf{1.0} & \textbf{67.34}±8.7 & \textbf{2.67}±4.4 & \textbf{1.0}                    & \textbf{74.19}±3.4 & \textbf{0.81}±0.23  & \textbf{1.0}                          \\ 
        \hline
        \multirow{3}{*}{\begin{tabular}[c]{@{}c@{}}Graph-\\SAGE\end{tabular}} & GD                      & 17.16±12.2          & 31.88±12.2         & 1.0          & 55.85±22.5         & 40.71±20.3         & 1.0          & 19.07±14.1         & 36.68±10.1        & 1.0                             & OOM                 & OOM            & 0                                     \\
                                                                              & ENN                     & 28.73±5.6           & 20.31±5.6          & 1.0          & 5.88±3.9           & 90.68±4.3          & 1.0          & ~8.14±8.6          & 47.61±7.6         & 1.0                             & OOM                 & OOM            & 0                                     \\
                                                                              & \egnn                   & \textbf{43.52}±10.8 & \textbf{5.12}±10.8 & \textbf{1.0} & \textbf{96.50}±0.1 & \textbf{~0.05}±0.1 & \textbf{1.0} & \textbf{67.91}±2.9 & \textbf{0.64}±2.3 & \textbf{1.0}                    & \textbf{76.27}±0.6 & \textbf{0.17}±0.10  & \textbf{1.0}                          \\
        \hline
        \end{tabular}}
        \end{table}

\begin{figure}[ht]
    \centering
    \includegraphics[width=0.24\linewidth]{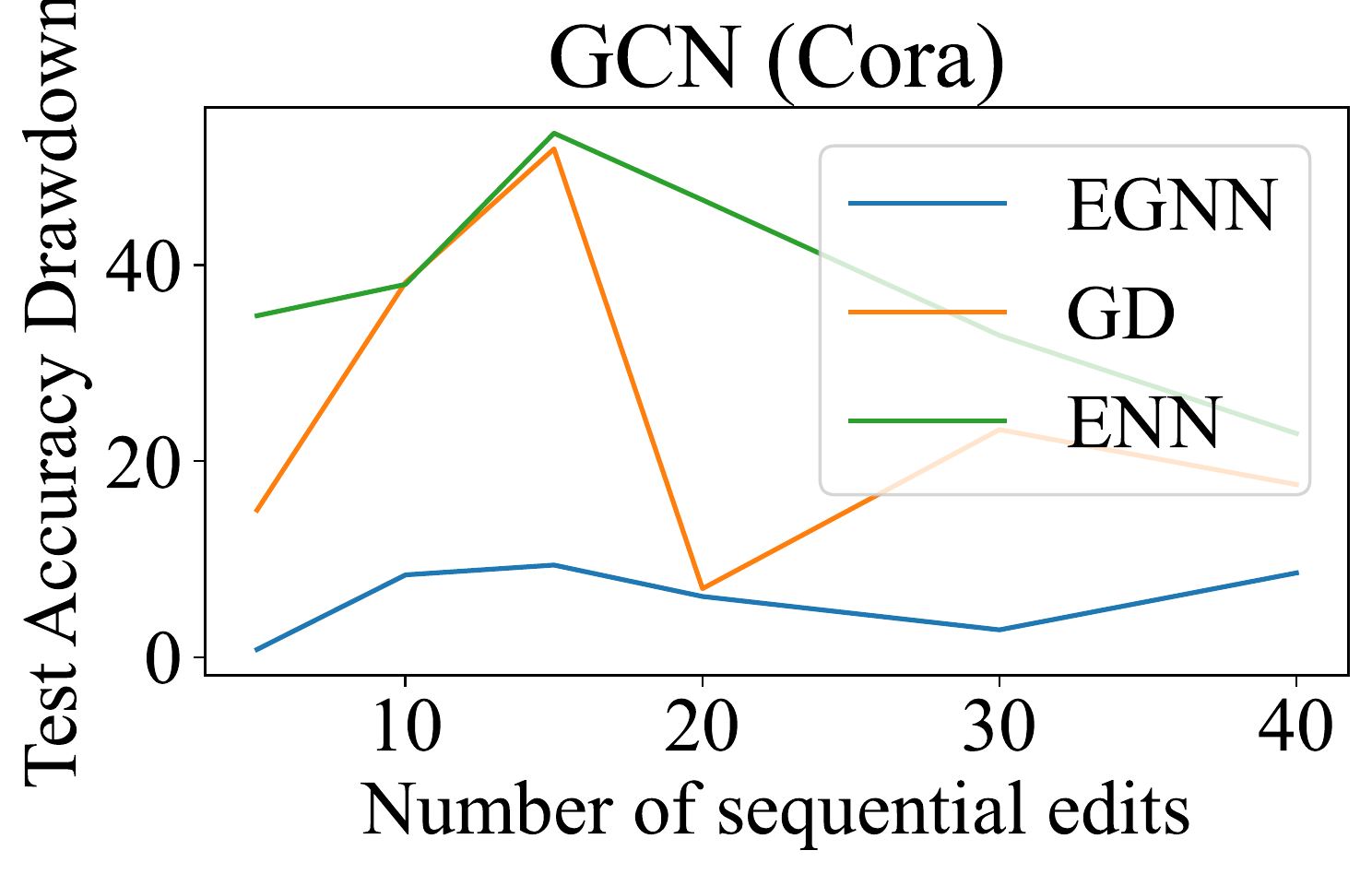}
    \includegraphics[width=0.24\linewidth]{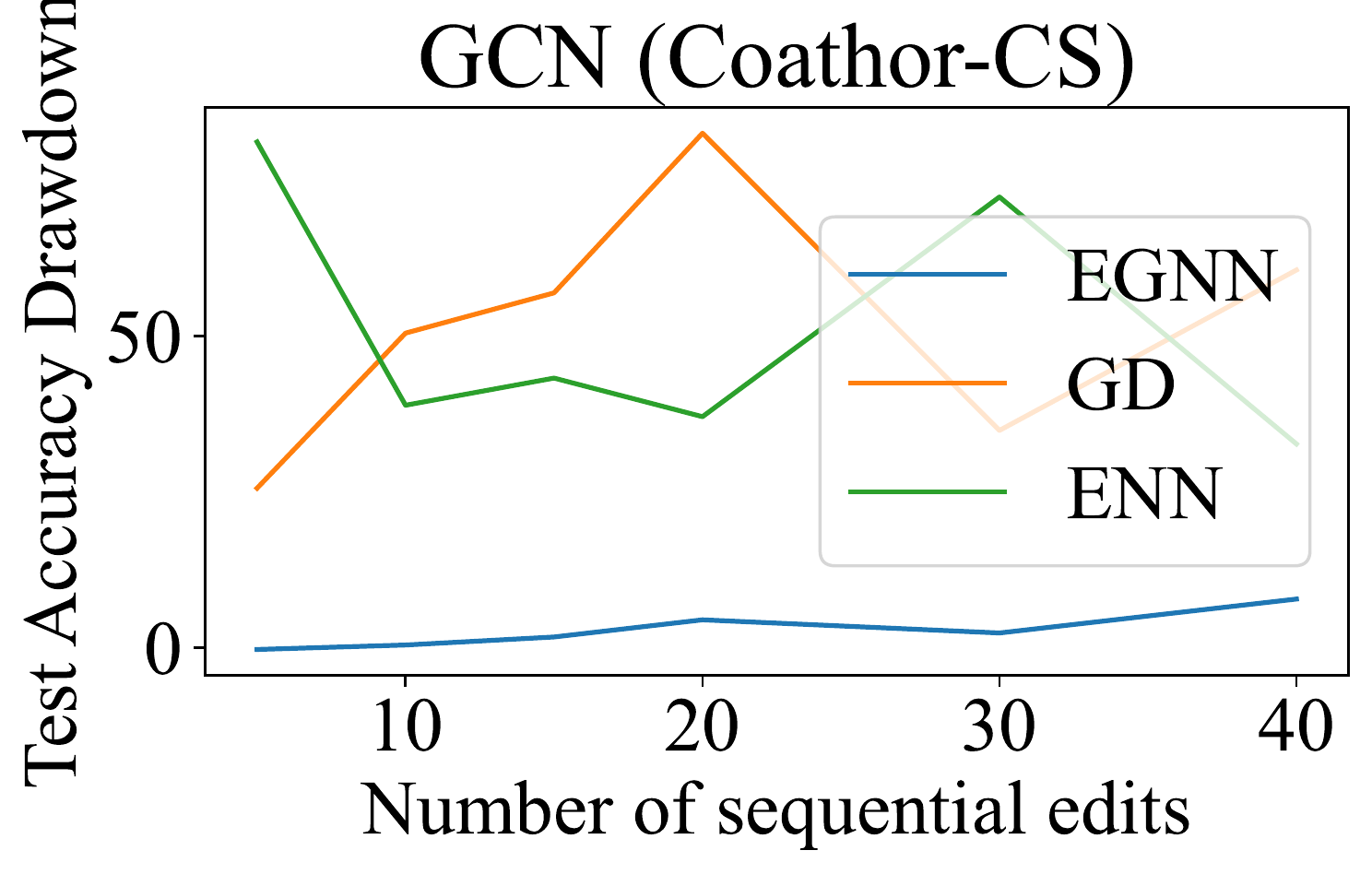}
    \includegraphics[width=0.24\linewidth]{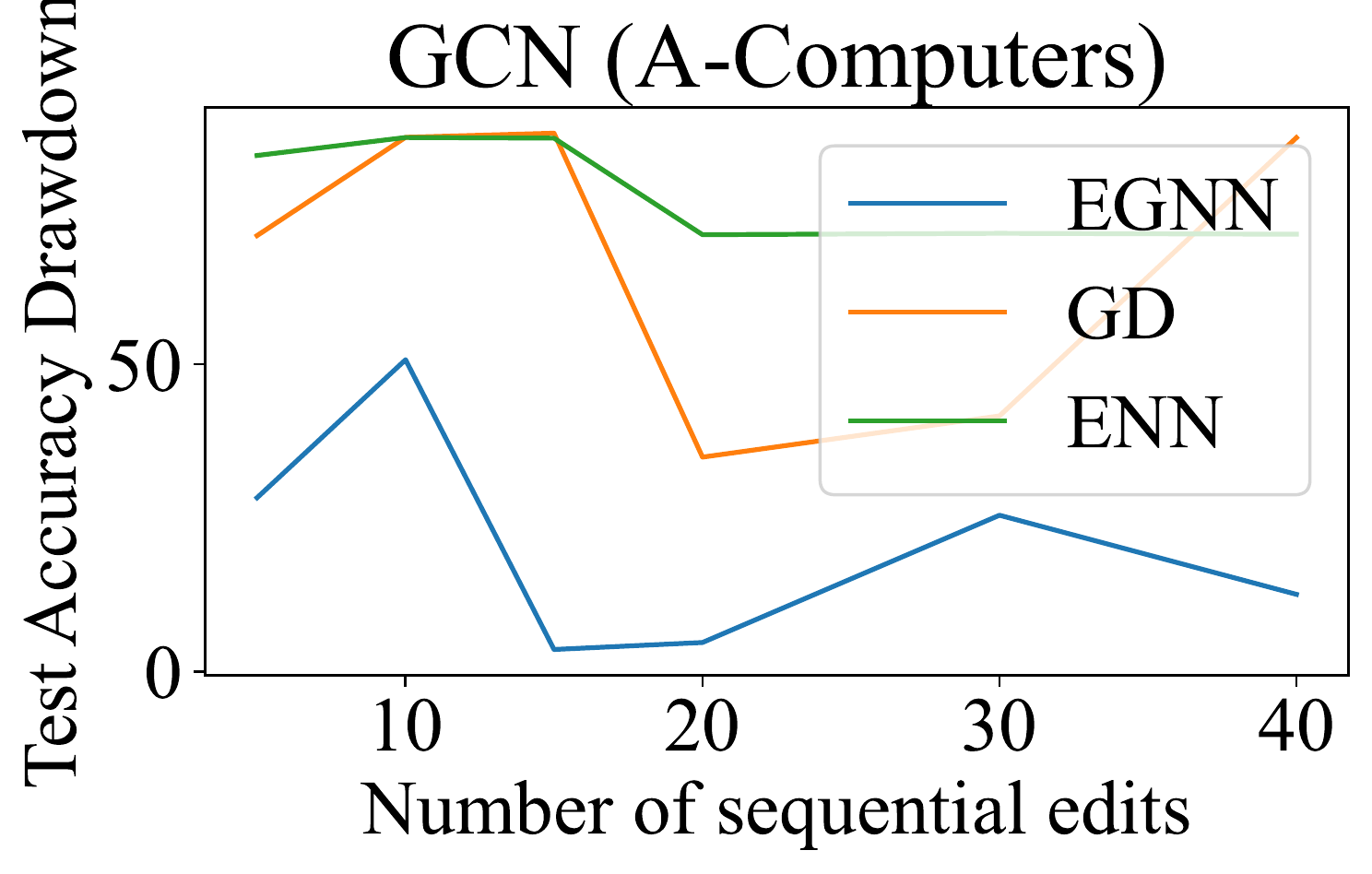}
        \includegraphics[width=0.24\linewidth]{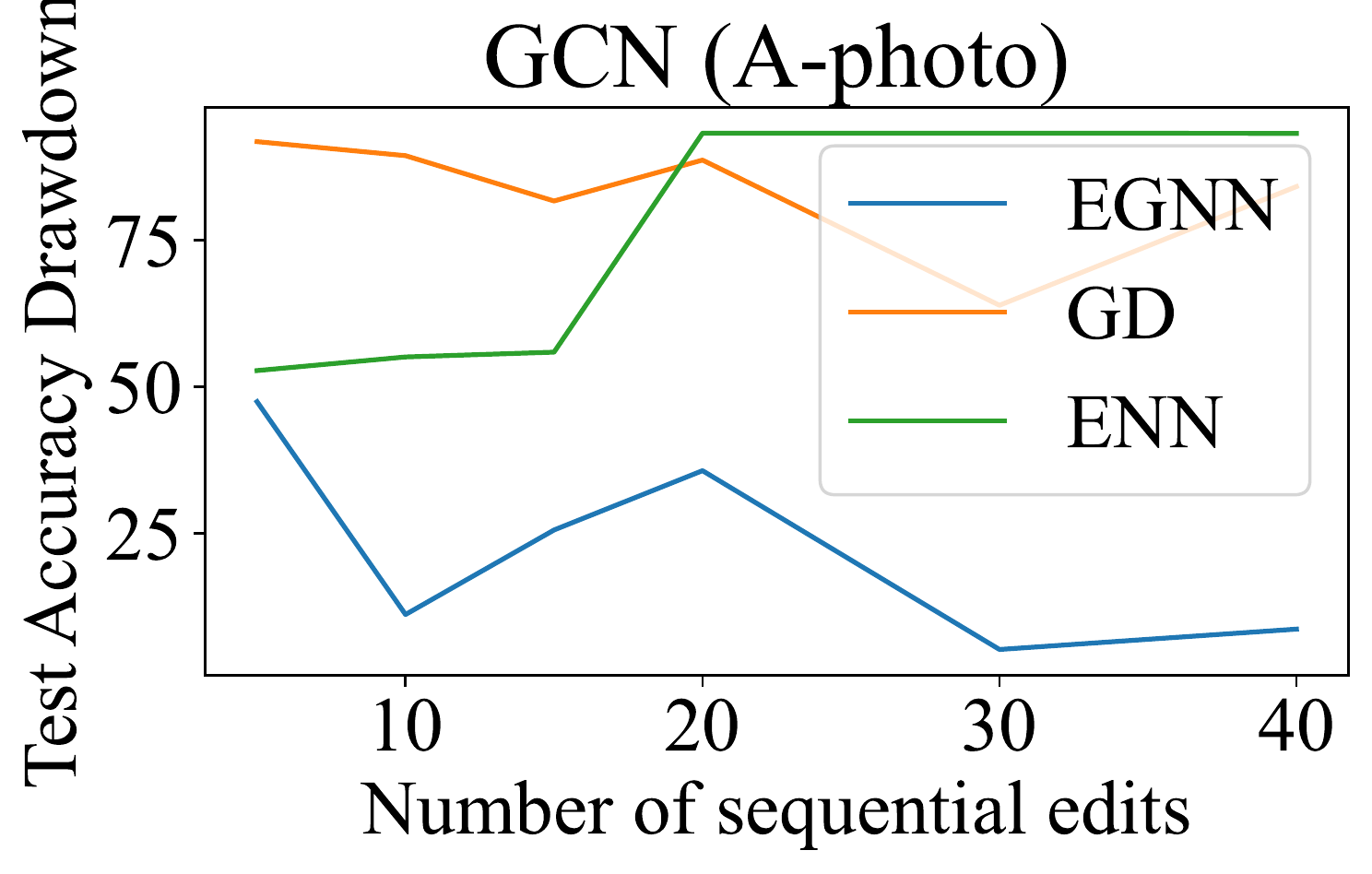}
    \caption{Sequential edit drawdown of GCN on four small scale datasets.}
    \label{fig:seq_gcn_1}
\end{figure}

\begin{figure}[ht]
    \centering
    \includegraphics[width=0.24\linewidth]{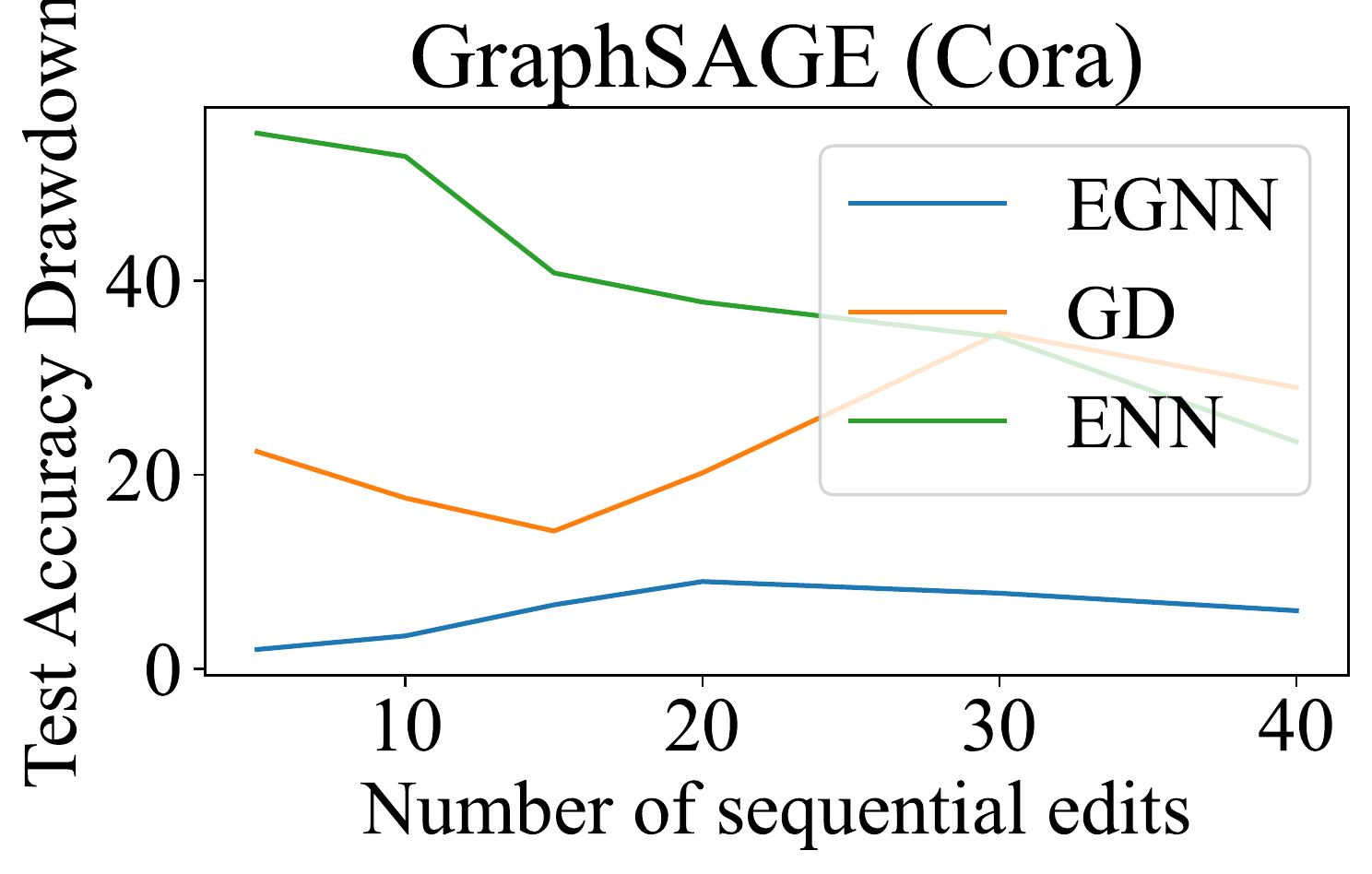}
    \includegraphics[width=0.24\linewidth]{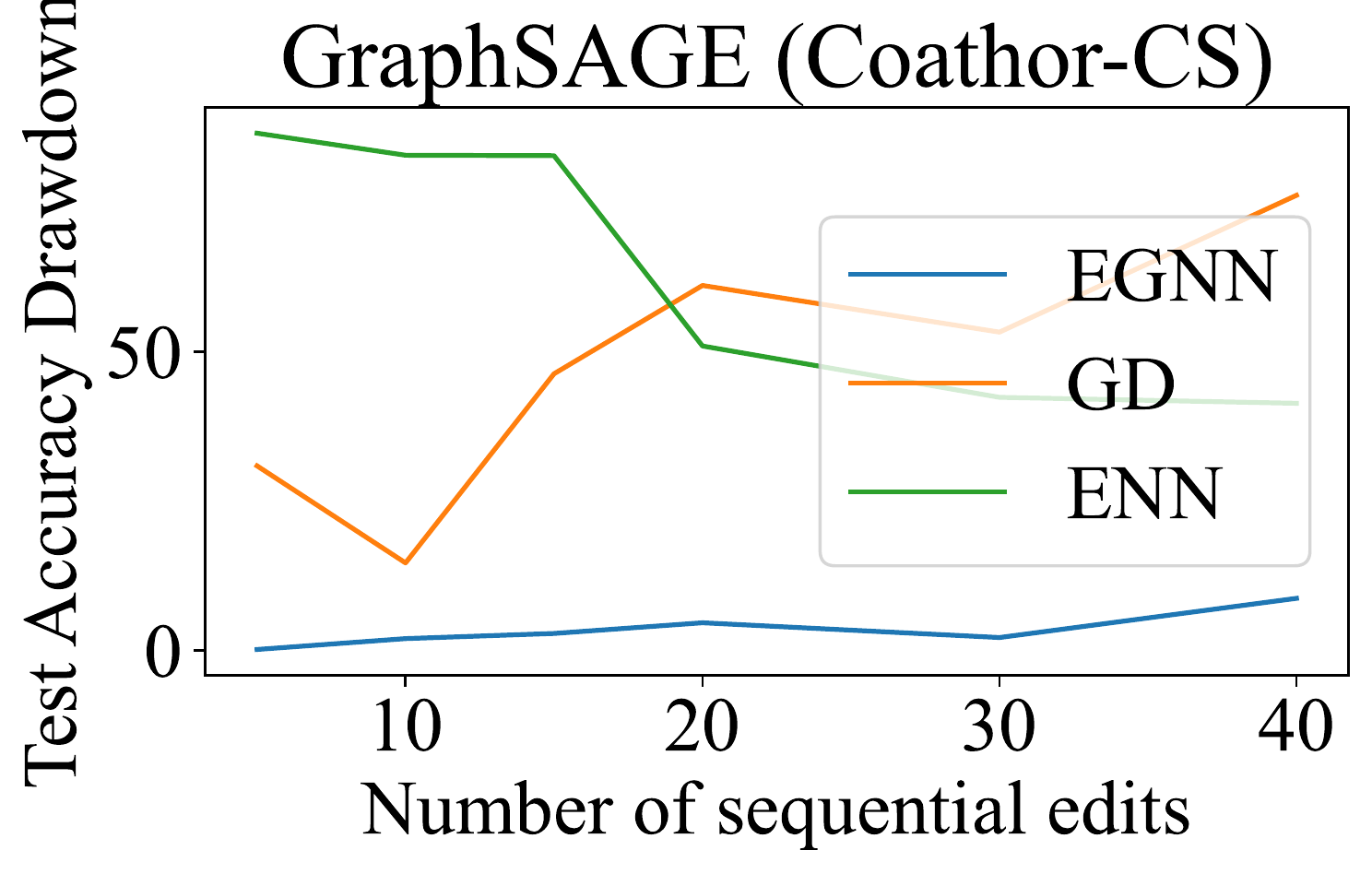}
    \includegraphics[width=0.24\linewidth]{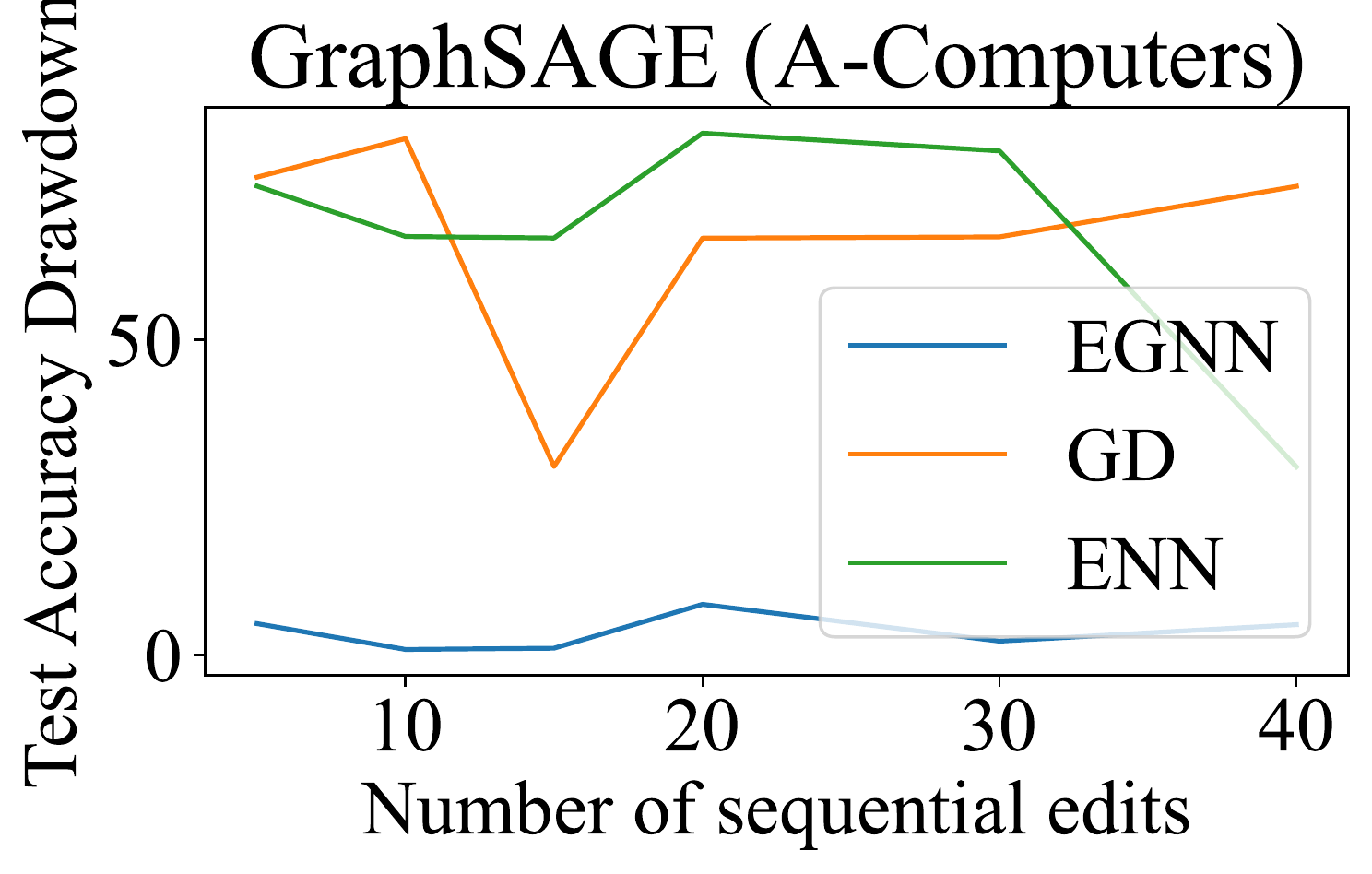}
        \includegraphics[width=0.24\linewidth]{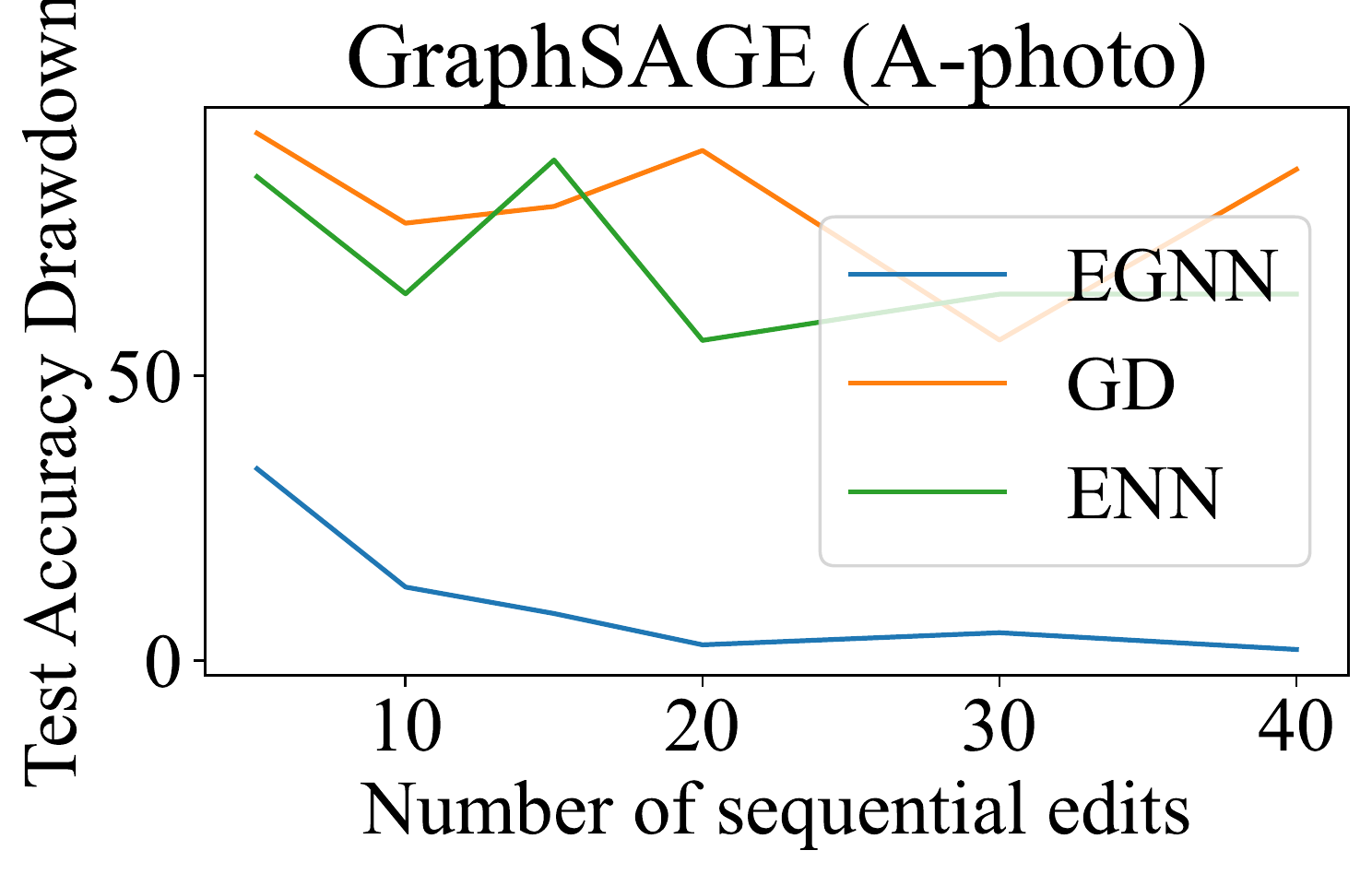}
    \caption{Sequential edit drawdown of GraphSAGE on four small scale datasets.}
    \label{fig:seq_sage_1}
\end{figure}

\begin{figure}[ht]
    \centering
    \includegraphics[width=0.24\linewidth]{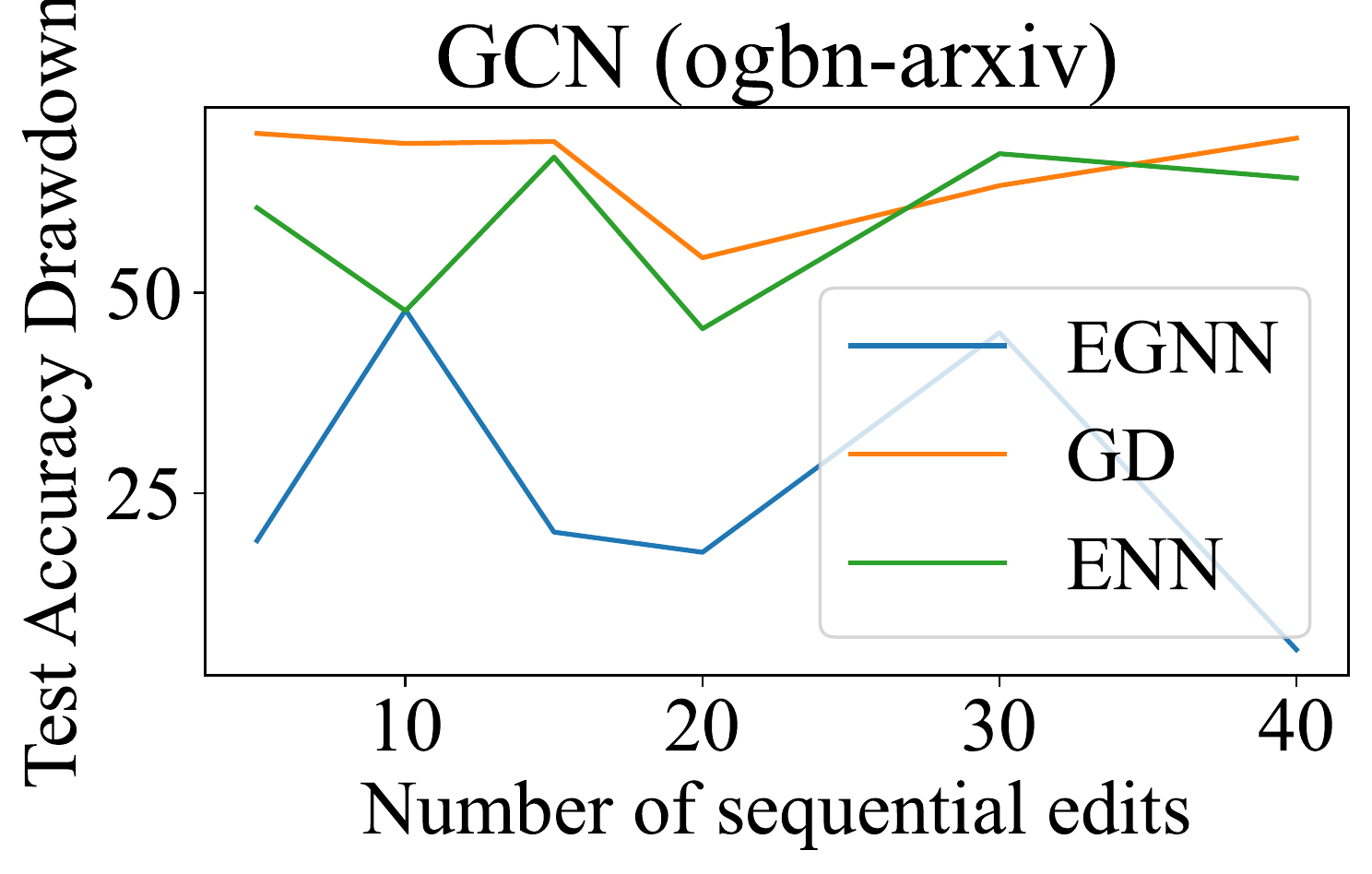}
    \includegraphics[width=0.24\linewidth]{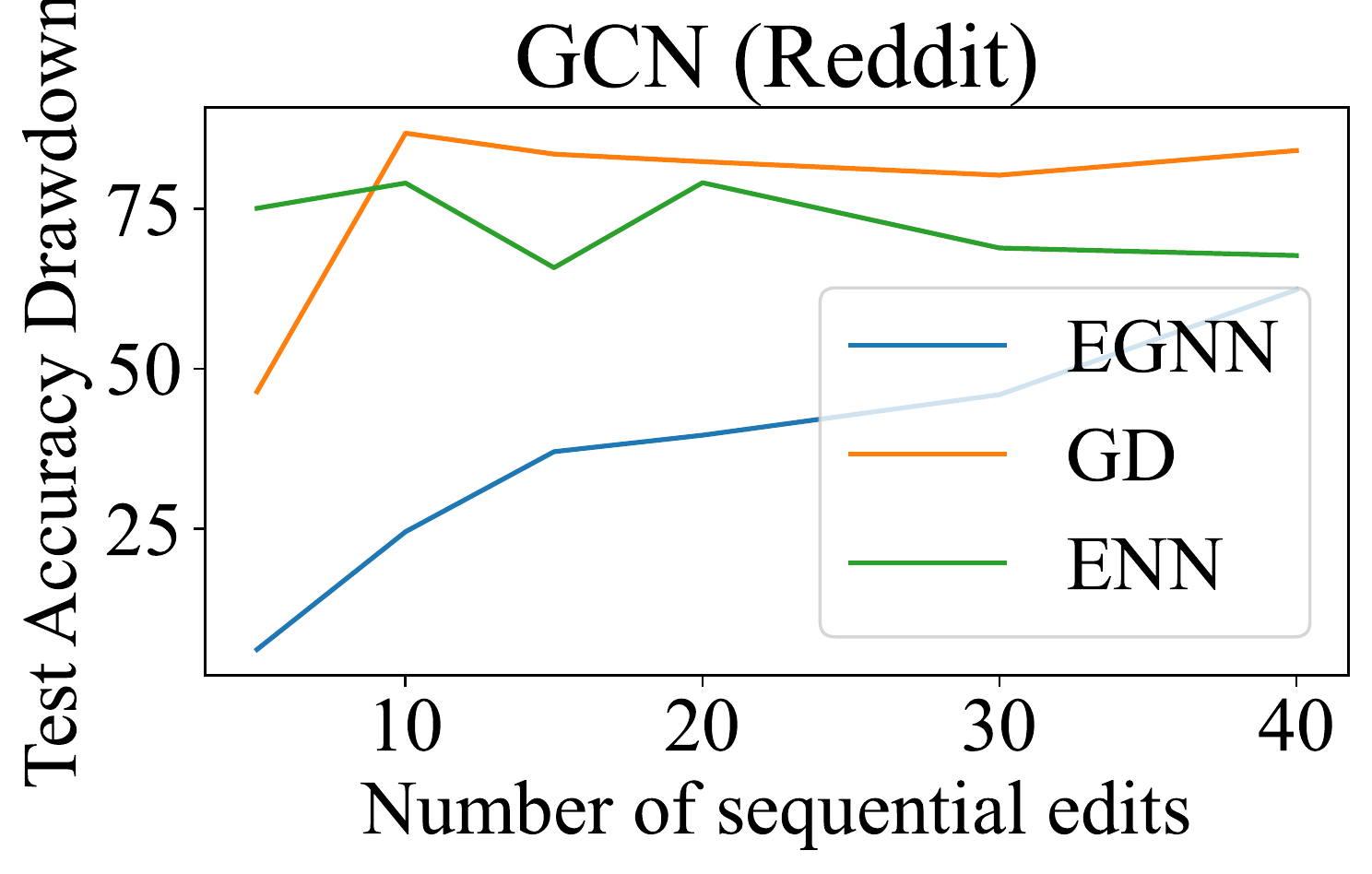}
    \includegraphics[width=0.24\linewidth]{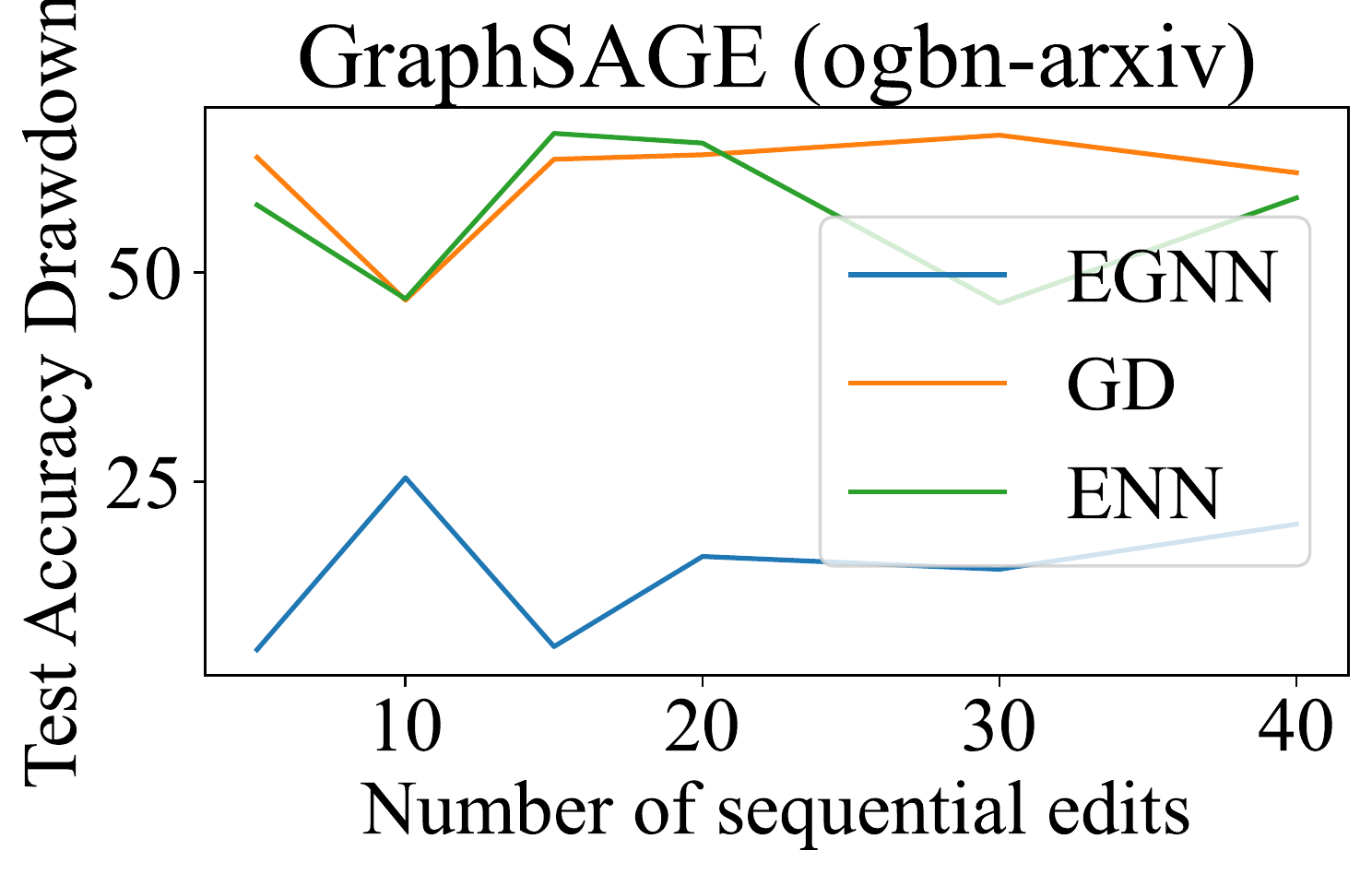}
    \includegraphics[width=0.24\linewidth]{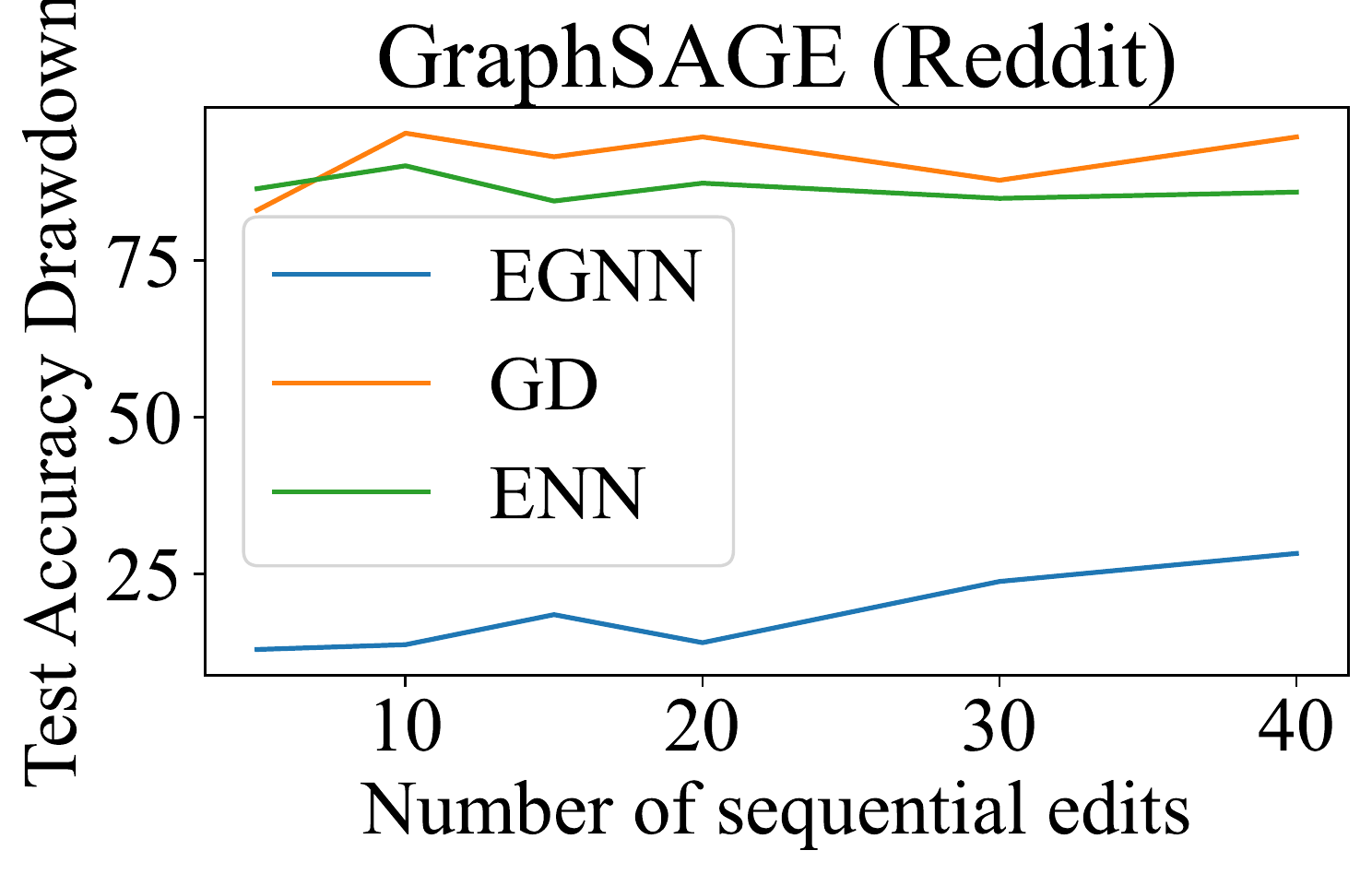}
    \caption{Sequential edit test drawdown of GCN and GraphSAGE on Reddit and ogbn-arxiv dataset.}
    \label{fig:seq_large_gcn_sage}
\end{figure}

\begin{figure}[ht!]
    \centering
    \begin{subfigure}[h]{0.25\linewidth}
      \includegraphics[width=1\linewidth]{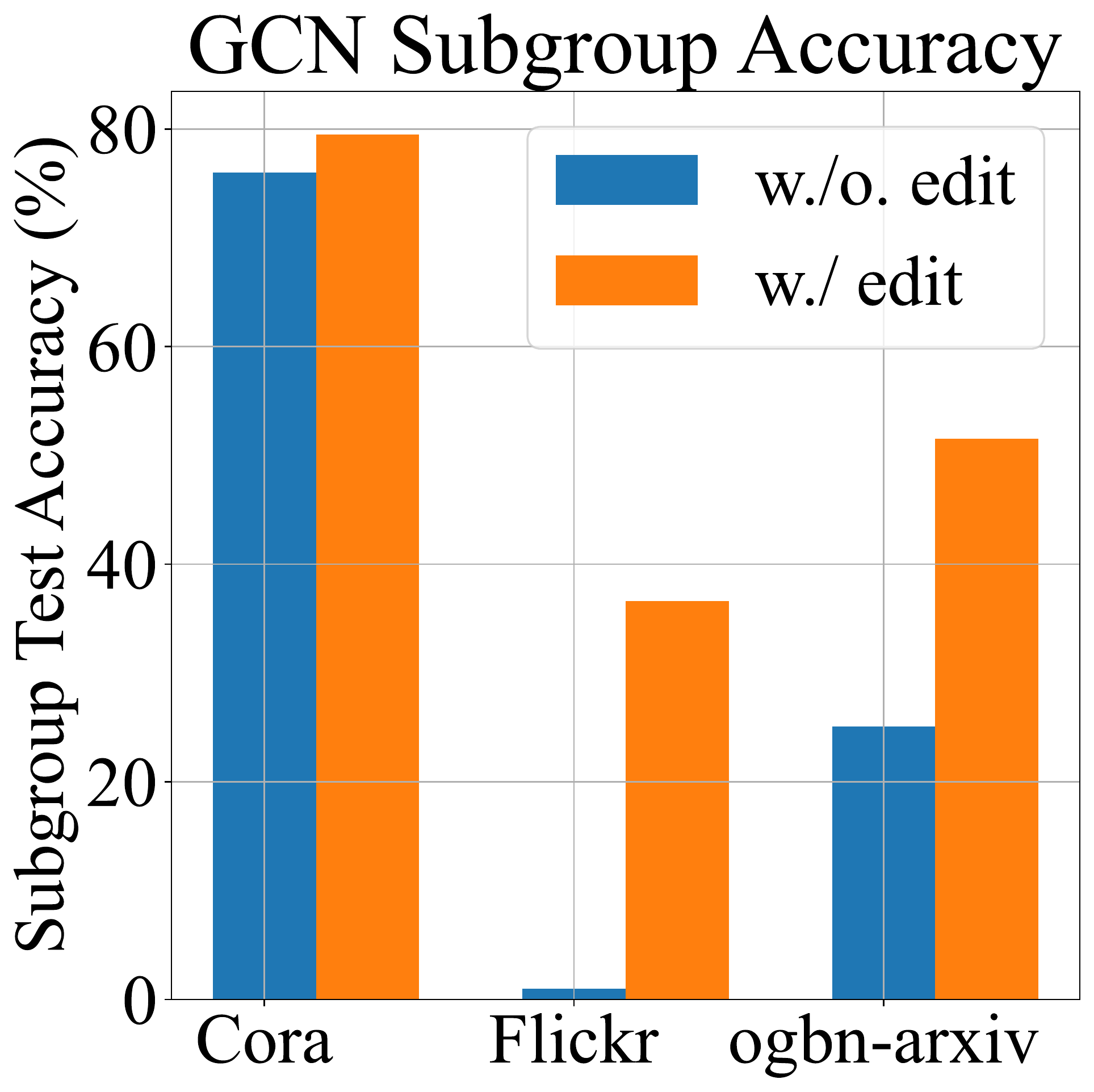}
      \caption{\small{GCN Subgroup Acc.}}
      \label{fig: GCN Subgroup Acc}
    \end{subfigure}%
    \begin{subfigure}[h]{0.25\linewidth}
      \includegraphics[width=1\linewidth]{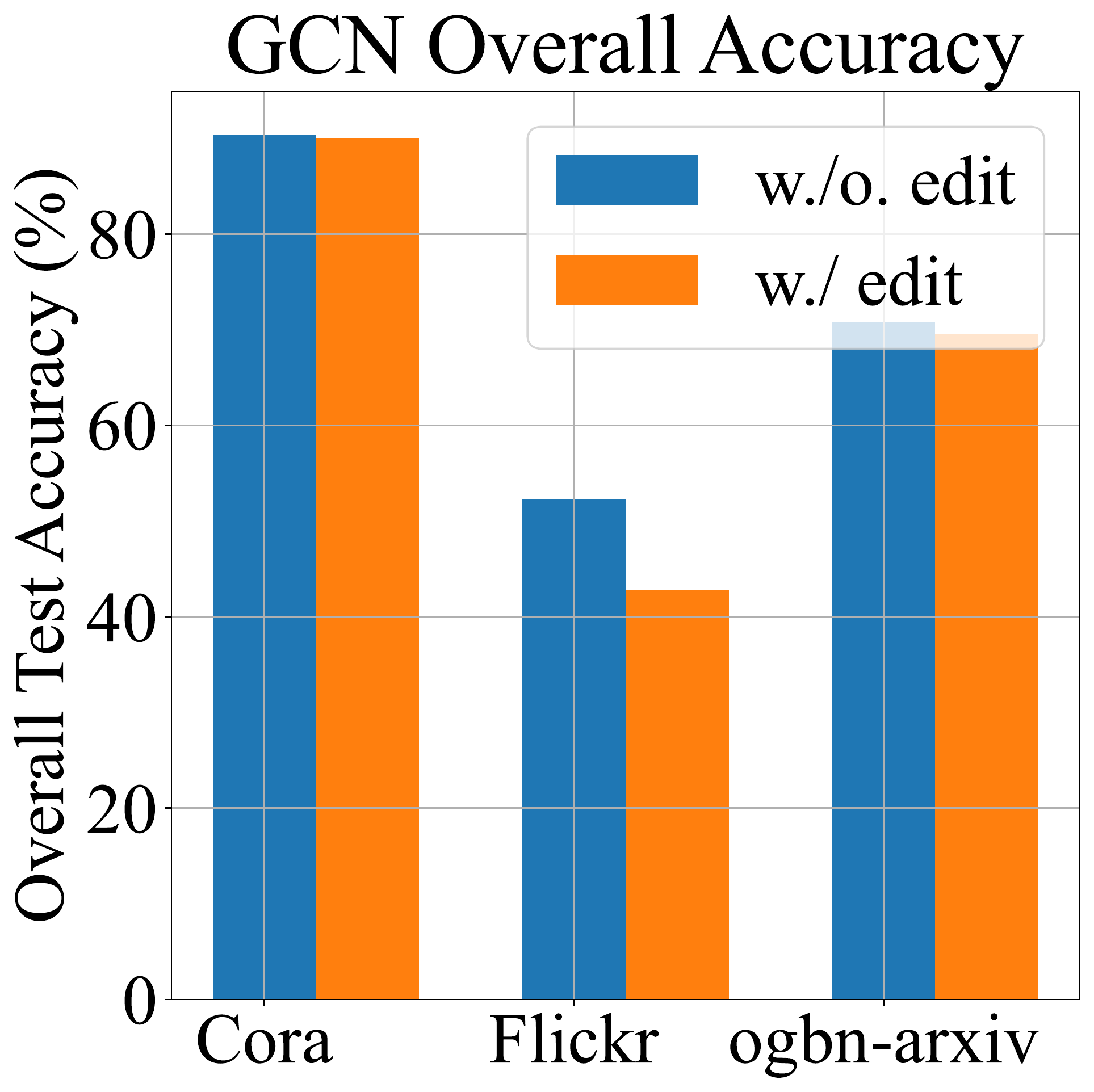}
        \caption{\small{GCN Overall Acc.}}
        \label{fig: GCN Overall Acc}
    \end{subfigure}%
        \begin{subfigure}[h]{0.25\linewidth}
      \includegraphics[width=1\linewidth]{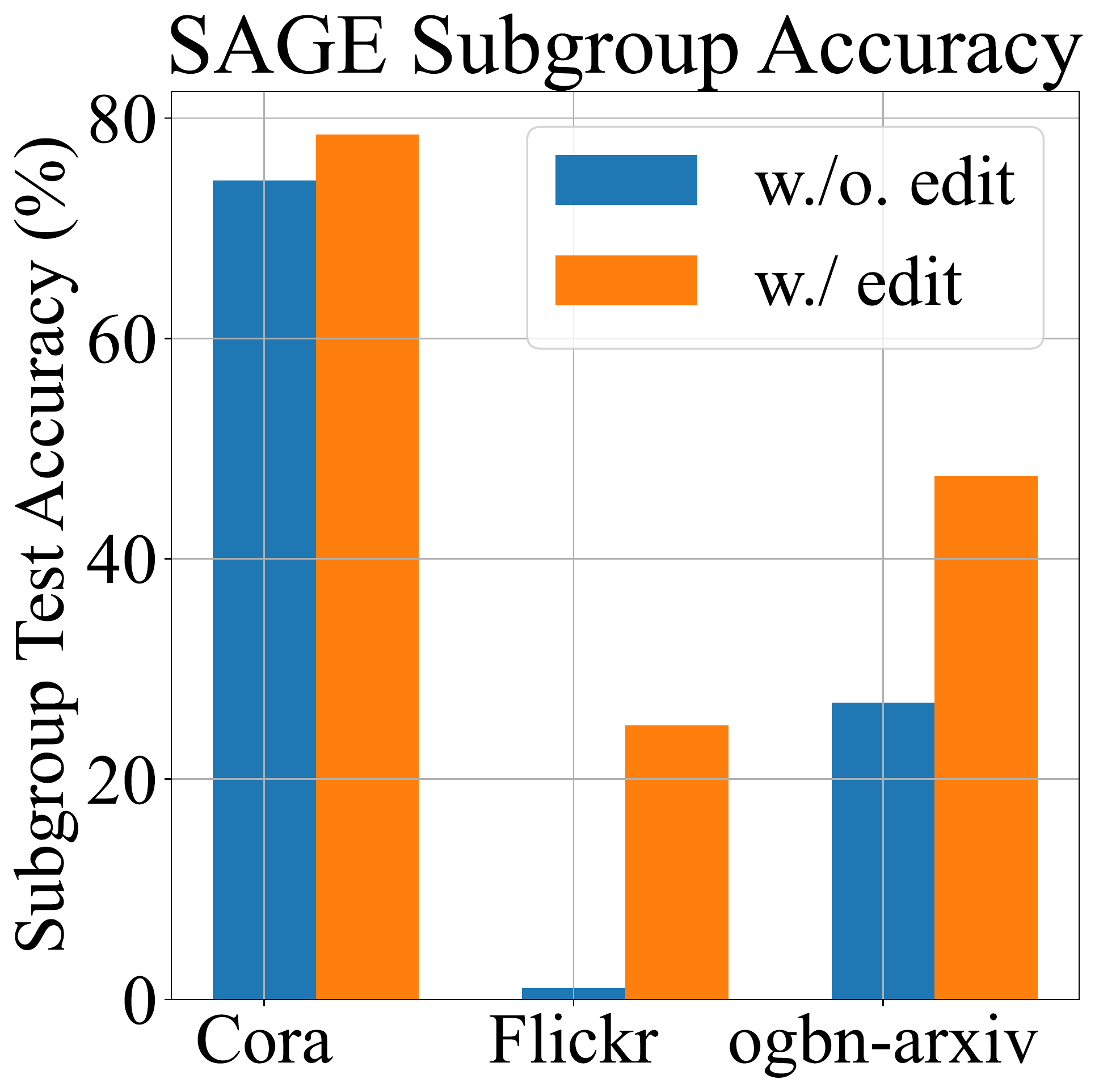}
        \caption{\small{SAGE Subgroup Acc.}}
        \label{fig: SAGE Subgroup Acc}
    \end{subfigure}%
        \begin{subfigure}[h]{0.25\linewidth}
      \includegraphics[width=1\linewidth]{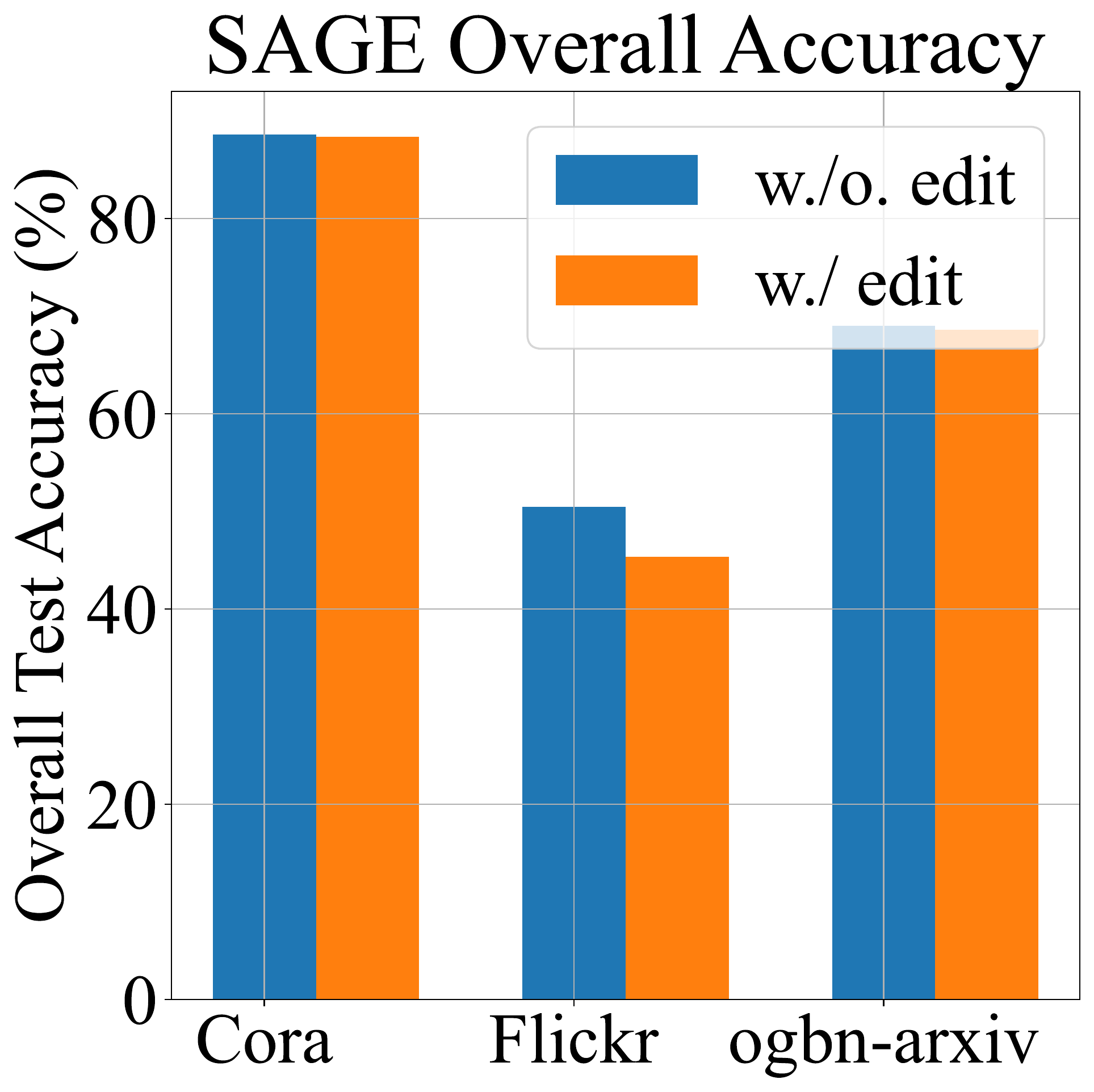}
    \caption{\small{SAGE Overall Acc.}}
    \label{fig: SAGE Overall Acc}
    \end{subfigure}%
    \caption{The subgroup and overall test accuracy before and after one single edit.
    The results are averaged over 50 independent edits.}
    \label{fig: generalization}
    \vspace{-1em}
\end{figure}


In many real-world applications, it is common to encounter situations where our trained model produces incorrect predictions on unseen data. 
It is crucial to address these errors as soon as they are identified. 
To assess the usage of editors in real-world applications ($\textbf{RQ1}$), 
\textbf{we select misclassified nodes from the validation set, which is not seen during the training process.}
Then we employ the editor to correct the model's predictions for those misclassified nodes, and measure the drawdown and edit success rate on the test set.

The results after editing on a single node are shown in Table \ref{tab: small_exp_res} and Table \ref{tab: main_exp_res}. 
We observe that 

\ding{182} \emph{Unlike editing Transformers on text data \citep{mitchell2021fast, mitchell2022memory, huang2023transformerpatcher}, all editors can successfully correct the model prediction in graph domain.}
As shown in Table \ref{tab: main_exp_res}, all editors have $100\%$ success rate when edit GNNs.
In contrast, for transformers, the edit success rate is often less than $50\%$ and drawdown is much smaller than GNNs \citep{mitchell2021fast, mitchell2022memory, huang2023transformerpatcher}.
This observation suggests that \textbf{unlike transformers, GNNs can be easily perturbed to produce correct predictions. 
However, at the cost of huge drawdown on other unrelated nodes.
Thus, the main challenge lies in maintaining the locality between predictions for unrelated nodes before and after editing.}
This observation aligns with our initial analysis, which highlighted the interconnected nature of nodes and the edit on a single node may propagate throughout the entire graph.

\ding{183} \emph{\egnn significantly outperforms both GD and ENN in terms of the test drawdown.}
This is mainly because both GD and ENN try to correct the model's predictions by updating the parameters of Graph Neural Networks (GNNs). 
This process inevitably relies on neighbor propagation.
In contrast, \egnn has much better test accuracy after editing.
Notably, for Reddit, the accuracy drop decreases from roughly $80\%$ to $\approx 1\%$, which is significantly better than the baseline.
This is because \egnn decouples the neighbor propagation with the editing process.
Interestingly, ENN is significantly worse than the vanilla editor, i.e., GD, when applied to GNNs.
As shown in Appendix \ref{app: more_exp_res}, we found that this discrepancy arises from the ENN training procedure, which significantly compromises the model's performance to prepare it for editing. 

In Figure \ref{fig:seq_gcn_1}, \ref{fig:seq_sage_1}, and \ref{fig:seq_large_gcn_sage} we present the ablation study under the sequential setting. 
This is a more challenging scenario where the model is edited sequentially as errors arise.
In particular, we plot the test accuracy drawdown against the number of sequential edits for GraphSAGE on the ogbn-arxiv dataset.
We observe that 
\ding{184} \emph{\egnn consistently surpasses both GD and ENN in the sequential setting.}
However, the drawdown is considerably greater than that in the single edit setting. For instance, \egnn exhibits a 0.64\% drawdown for GraphSAGE on the ogbn-arxiv dataset in the single edit setting, which escalates up to a 20\% drawdown in the sequential edit setting.
These results also highlight the hardness of maintaining the locality of GNN prediction after editing.

\begin{wrapfigure}{r}{0.55\textwidth}
\vspace{-1.5em}
    \begin{subfigure}[h]{0.54\linewidth}
      \includegraphics[width=1\linewidth]{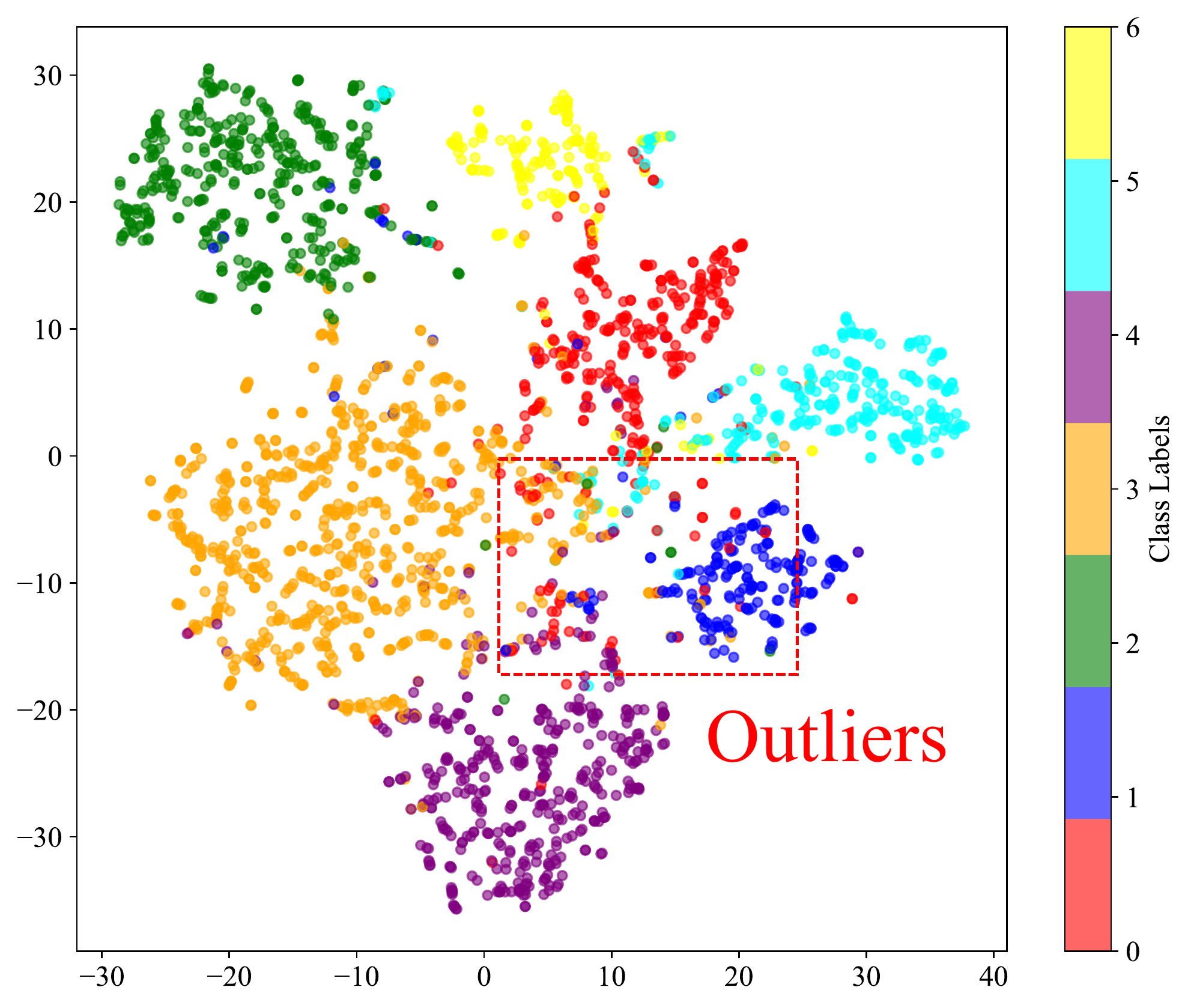}
      \caption{Before Edit}
    \end{subfigure}%
    \begin{subfigure}[h]{0.54\linewidth}
      \includegraphics[width=1\linewidth]
      {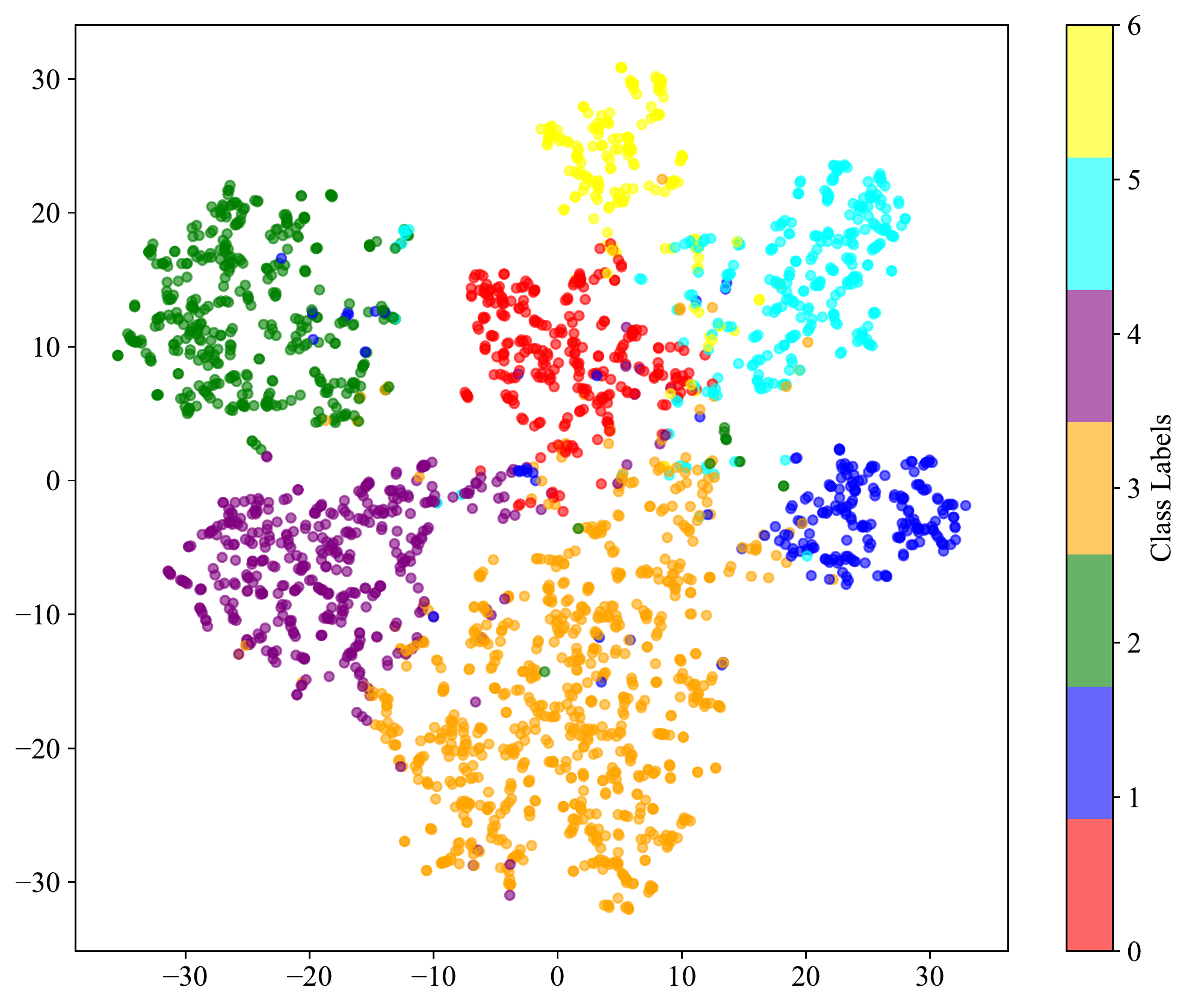}
            \caption{After Edit}
    \end{subfigure}%
    \caption{T-SNE visualizations of GNN embeddings before and after edits on the Cora dataset. The flipped nodes are all from class 0, which is marked in red color.}
    \label{fig: tsne}
\end{wrapfigure}

\subsection{The Generalization of the Edits of $\egnn$}

Ideally, we aim for the edit applied to a specific node to generalize to similar nodes while preserving the model's initial behavior for unrelated nodes. To evaluate the generalization of the \egnn edits, we conduct the following experiment:

\textbf{(1)} We first select a particular group (i.e., class) of nodes based on their labels.
\textbf{(2)} Next, we randomly flip the labels of $10\%$ of the training nodes within this group and train a GNN on the modified training set.
\textbf{(3)} For each flipped training node, we correct the trained model's prediction for that node back to its original class and assess whether the model's predictions for other nodes in the same group are also corrected.
If the model's predictions for other nodes in the same class are also corrected after modifying a single flipped node, it indicates that the \egnn edits can effectively generalize to address similar erroneous behavior in the model.

To answer \textbf{RQ2}, we conduct the above experiments and report the \textbf{subgroup and overall test accuracy} after performing a single edit on the flipped training node.
The results are shown in Figure \ref{fig: generalization}.
We observe that:
\ding{185} \emph{From Figure \ref{fig: GCN Subgroup Acc} and Figure \ref{fig: SAGE Subgroup Acc}, \egnn significantly improves the subgroup accuracy after performing even a single edit.} 
Notably, the subgroup accuracy is significantly lower than the overall accuracy.
For example, on Flickr dataset, both GCN and GraphSAGE have a subgroup accuracy of less than $5\%$ before editing.
This is mainly because the GNN is trained on the graph where $10\%$ labels of the training node in the subgroup are flipped.
However, even after editing on a single node, the subgroup accuracy is significantly boosted.
These results indicate that the \egnn edits can effectively generalize to address the wrong prediction on other nodes in the same group.
In Figure \ref{fig: tsne}, we also visualize the node embeddings before and after editing by \egnn on the Cora dataset.
We note that all of the flipped nodes are from class 0, which is marked in red color in Figure \ref{fig: tsne}.
Before editing, the red cluster has many outliers that lie in the embedding space of other classes. 
This is mainly because the labels of some of the nodes in this class are flipped.
In contrast, after editing, the nodes in the red cluster become significantly closer to each other, with a substantial reduction in the number of outliers.

\subsection{The Efficiency of $\egnn$}
We want to patch the model as soon as possible to correct errors as they appear.
Thus ideally, the editor should be efficient and scalable to large graphs.
Here we summarize the edit time and memory required for performing the edits in Table \ref{tab: exp_eff}.
We observe that \egnn is about $2\sim 5\times$ faster than the GD editor in terms of the wall-clock edit time.
This is because \egnn only updates the parameters of MLP, and totally gets rid of the 
expensive graph-based sparse operations \citep{exact, liu2022rsc, han2023mlpinit}.
Also, updating the parameters of GNNs requires storing the node embeddings in memory, which is directly proportional to the number of nodes in the graph and can be exceedingly expensive for large graphs
However, with \egnn, we only use node features for updating MLPs, meaning that memory consumption is not dependent on the graph size.
Consequently, \egnn can efficiently scale up to handle graphs with millions of nodes, e.g., ogbn-products, whereas the vanilla editor raises an OOM error.

\begin{table}
    \centering
    \captionsetup{skip=5pt} 
    \caption{The edit time and memory required for editing.}
    \label{tab: exp_eff}
    \resizebox{\textwidth}{!}{
    \begin{tabular}{cccccccccc} 
    \hline
    \multirow{2}{*}{}                                                     & \multirow{2}{*}{Editor} & \multicolumn{2}{c}{Flickr}                                                                                          & \multicolumn{2}{c}{Reddit}                                                                                          & \multicolumn{2}{c}{\begin{tabular}[c]{@{}c@{}}ogbn-\\arxiv\end{tabular}}                                            & \multicolumn{2}{c}{\begin{tabular}[c]{@{}c@{}}ogbn-\\products\end{tabular}}                                          \\ 
    \cline{3-10}
                                                                          &                         & \begin{tabular}[c]{@{}c@{}}Edit\\Time (ms)\end{tabular} & \begin{tabular}[c]{@{}c@{}}Peak\\Memory (MB)\end{tabular} & \begin{tabular}[c]{@{}c@{}}Edit\\Time (ms)\end{tabular} & \begin{tabular}[c]{@{}c@{}}Peak\\Memory (MB)\end{tabular} & \begin{tabular}[c]{@{}c@{}}Edit\\Time (ms)\end{tabular} & \begin{tabular}[c]{@{}c@{}}Peak\\Memory (MB)\end{tabular} & \begin{tabular}[c]{@{}c@{}}Edit\\Time (ms)\end{tabular} & \begin{tabular}[c]{@{}c@{}}Peak\\Memory (MB)\end{tabular}  \\ 
    \hline
    \multirow{2}{*}{GCN}                                                  & GD                      & 379.86                                                  & 707                                                       & 1835.24                                                 & 3429                                                      & 663.17                                                  & 967                                                       & OOM                                                     & OOM                                                        \\
                                                                          & \egnn                   & 246.63                                                  & 315                                                       & 765.15                                                  & 2089                                                      & 299.71                                                  & 248                                                       & 5122.53                                                 & 5747                                                       \\ 
    \hline
    \multirow{2}{*}{\begin{tabular}[c]{@{}c@{}}Graph-\\SAGE\end{tabular}} & GD                      & 712.07                                                  & 986                                                       & 4781.92                                                 & 5057                                                      & 668.77                                                  & 1109                                                      & OOM                                                     & OOM                                                        \\
                                                                          & \egnn                   & 389.37                                                  & 328                                                       & 1516.68                                                 & 2252                                                      & 174.82                                                  & 260                                                       & 5889.59                                                 & 6223                                                       \\
    \hline
    \end{tabular}}
    \end{table}
\section{Conclusion}

In this paper, we explore a and important problem, i.g., GNNs model editing for node classification. 
We first empirically observe that the vanilla model editing method may not perform well due to node aggregation, and then theoretically investigate the underlying reason through the lens of locality loss landscape with quantitative analysis. Furthermore, we propose EGNN to correct misclassified samples while preserving other intact nodes, via stitching a trainable MLP. 
In this way, the power of GNNs for prediction and the editing-friendly
MLP can be integrated together in EGNN.

\bibliographystyle{plainnat}
\bibliography{ref}

\clearpage
\appendix
\section{Experimental Setting}
\label{app: exp_setting}

\subsection{Datasets for node classification}\label{sec:app:setting:data}
The details of datasets used for node classification are listed as follows:
\begin{itemize}
\item Cora \citep{sen2008collective} is the citation network.
The dataset contains 2,708 publications with 5,429 links, and each publication is described by a 1,433-dimensional binary vector, indicating the presence or absence of corresponding words from a fixed vocabulary.

\item A-computers \citep{shchur2018pitfalls} is the segment of the Amazon co-purchase graph, where nodes represent goods, edges indicate that two goods are frequently bought together, node features are bag-of-words encoded product reviews.

\item A-photo \citep{shchur2018pitfalls}  is similar to A-computers, which is also the segment of the Amazon co-purchase graph, where nodes represent goods, edges indicate that two goods are frequently bought together, node features are bag-of-words encoded product reviews.

\item Coauthor-CS \citep{shchur2018pitfalls} is the co-authorship graph based on the Microsoft Academic Graph from the KDD Cup 2016 challenge 3. 
Here, nodes are authors, that are connected by an edge if they
co-authored a paper; node features represent paper keywords for each author’s papers, and class labels indicate most active fields of study for each author.

\item Reddit \citep{gsage} is constructed by Reddit posts. The node in this dataset is a post belonging to different communities.
\item \textit{ogbn-arxiv} \citep{ogb} is the citation network between all arXiv papers. Each node denotes a paper and each edge denotes citation between two papers. The node features are the average 128-dimensional word vector of its title and abstract.
\item \textit{ogbn-prducts} \citep{ogb} is Amazon product co-purchasing network. Nodes represent products in Amazon, and edges between two products indicate that the products are purchased together. 
Node features are low-dimensional representations of the product description text.
\end{itemize}

\begin{table}[h!]
\centering
\caption{Statistics for datasets used for node classification.}
\label{tab:dataset:nc}
\begin{tabular}{crrrrr} 
\toprule
Dataset                & \# Nodes. & \# Edges   & \# Classes & \# Feat & Density  \\ 
\midrule
Cora                   & 2,485     & 5,069      & 7          & 1433    & 0.72\textperthousand     \\
A-computers            & 13,381    & 245,778    & 10         & 767     & 2.6\textperthousand      \\
A-photo                & 7,487     & 119,       & 8          & 745     & 4.07\textperthousand     \\
Coauthor-CS            & 18,333    & 81,894     & 15         & 6805    & 0.49\textperthousand     \\
Flickr                 & 89,250    & 899,756    & 7          & 500     & 0.11\textperthousand     \\
Reddit                 & 232,965   & 23,213,838 & 41         & 602     & 0.43\textperthousand     \\
\textit{ogbn-arxiv}    & 169,343   & 1,166,243  & 40         & 128     & 0.04\textperthousand     \\
\textit{ogbn-products} & 2,449,029 & 61,859,140 & 47         & 218     & 0.01\textperthousand     \\
\bottomrule
\end{tabular}
\end{table}


\subsection{Baselines for node classification}\label{sec:app:setting:base}
We present the details of the hyperparameters of GCN, GraphSAGE, and the stitched MLP modules in Table \ref{tab: config_nc}.
We use the Adam optimizer for all these models.

\begin{table}[t]
    \caption{Training configuration for employed models}\label{tab: config_nc}
    \centering
    \setlength{\tabcolsep}{2pt}
    \begin{tabular}{l|crrrrrrr}
        \toprule
        Model                                  & Dataset        & \#Layers & \#Hidden & Learning rate & Dropout  & Epoch \\
        \midrule
        \multirow{8}{*}{\rotatebox{90}{GraphSAGE}} & Cora          & $2$      & $32$    & $0.01$      & $0.1$    & $200$ \\
                                                 & A-computers    & $2$      & $32$    & $0.01$      & $0.1$     & $400$ \\
                                                 & A-photo        & $2$      & $32$    & $0.01$      & $0.1$    & $400$ \\
                                                 & Coauthor-CS    & $2$      & $32$    & $0.01$       & $0.1$    & $400$ \\
                                                 & Flickr         & $2$      & $256$    & $0.01$       & $0.3$     & $400$ \\
                                                 & Reddit         & $2$      & $256$    & $0.01$       & $0.5$     & $400$ \\
                                                 & ogbn-arxiv     & $3$      & $128$    & $0.01$       & $0.5$         & $500$ \\
                                                 & ogbn-products  & $3$      & $256$    & $0.002$       & $0.5$      & $500$ \\
        \midrule
        \multirow{8}{*}{\rotatebox{90}{GCN}}      & Cora          & $2$      & $32$    & $0.01$      & $0.1$      & $200$ \\
                                                 & A-computers    & $2$      & $32$    & $0.01$      & $0.1$          & $400$ \\
                                                 & A-photo        & $2$      & $32$    & $0.01$      & $0.1$      & $400$ \\
                                                 & Coauthor-CS    & $4$      & $32$    & $0.01$       & $0.1$          & $400$ \\
                                                 & Flickr         & $2$      & $256$    & $0.01$       & $0.3$       & $400$ \\
                                                & Reddit        & $2$      & $256$    & $0.01$       & $0.5$           & $400$ \\
                                                & ogbn-arxiv    & $3$      & $128$    & $0.01$       & $0.5$           & $500$ \\
                                                & ogbn-products & $3$      & $256$    & $0.002$       & $0.5$   & $500$ \\
        \midrule
        \multirow{8}{*}{\rotatebox{90}{MLP}}      & Cora          & $2$      & $32$    & $0.01$      & $0.1$      & $200$ \\
                                                 & A-computers    & $2$      & $32$    & $0.01$      & $0.1$          & $400$ \\
                                                 & A-photo        & $2$      & $32$    & $0.01$      & $0.1$      & $400$ \\
                                                 & Coauthor-CS    & $4$      & $32$    & $0.01$       & $0.1$          & $400$ \\
                                                 & Flickr         & $2$      & $256$    & $0.01$       & $0.3$       & $400$ \\
                                                & Reddit        & $2$      & $256$    & $0.01$       & $0.5$           & $400$ \\
                                                & ogbn-arxiv    & $3$      & $128$    & $0.01$       & $0.5$           & $500$ \\
                                                & ogbn-products & $3$      & $256$    & $0.002$       & $0.5$   & $500$ \\
        \bottomrule
    \end{tabular}
\end{table}

\subsection{Hardware and software configuration}\label{sec:app:soft_hard}

All experiments are executed on a server with 500GB main memory, two AMD EPYC 7513 CPUs.
All experiments are done with a single NVIDIA RTX A5000 (24GB).
The software and package version is specified in Table \ref{tab: package config}:

\begin{table}[h!]
\centering
\captionsetup{skip=5pt} 
\caption{Package configurations of our experiments.}
\label{tab: package config}
\begin{tabular}{ccc} 
\hline
Package & Version \\
\hline
CUDA & 11.3 \\
pytorch          &          1.10.2 \\
torch-geometric    &           1.7.2 \\
torch-scatter       &          2.0.8 \\
torch-sparse         &         0.6.12 \\
\hline
\end{tabular}
\end{table}

\section{Limitations and Future Work}
\label{app:limit}

Despite that \egnn is effective, generalized, and efficient, its main limitation is that currently it will incur the larger inference latency, due to the extra MLP module.
However, we note that this inference overhead is negligible. 
This is mainly because the computation of MLP only involve dense matrix operation, which is way more faster than the graph-based sparse operation. The future work comprises several research directions, including (1) Enhancing the efficiency of editable graph neural networks training through various perspectives (e.g., model initialization, data, and gradient); (2) understanding why vanilla editable graph neural networks training fails from other perspectives (e.g., interpretation and information bottleneck) \citep{lundberg2017unified,tishby2000information, pid}; (3)
Advancing the scalability, speed, and memory efficiency of editable graph neural networks training \citep{exact, liu2022rsc, han2023mlpinit}; (4) Expanding the scope of editable training for other tasks (e.g., link prediction, and knowledge graph) \citep{lu2011link,wang2014knowledge}; (5) Investigating the potential issue concerning privacy, robustness, and fairness in the context of editable graph neural networks training \citep{zheleva2008preserving,jiang2023weight,jin2020graph,dai2021say,jiang2022generalized,han2023retiring, DBLP:journals/corr/abs-2006-08315}.


\section{More Experimental Results}
\label{app: more_exp_res}

\subsection{More Loss Landscape Results}

\begin{figure}[h!]
    \centering
    \includegraphics[width=0.99\linewidth]{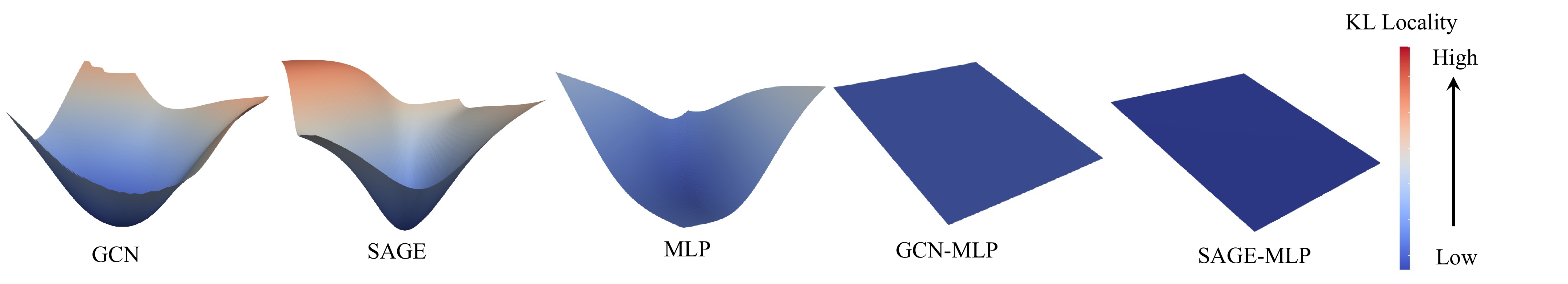}
    \caption{The loss landscape of various GNNs architectures on Flickr dataset.}
    \label{fig: landscape_2}
\end{figure}

We visualize the locality loss landscape for Flickr dataset in Figure~\ref{fig: landscape_2}. Similarly, $Z$ axis denotes the KL divergence, X-Y axis is centered on the original model weights before editing and quantifies the weight perturbation scale after model editing. We observe similar observations: (1) GNNs architectures (e.g., GCN and GraphSAGE) suffer from a much sharper loss landscape at the convergence of original model weights. KL divergence locality loss is dramatically enhanced even for slight weights editing. 
(2) MLP shows a flatter loss landscape and demonstrates mild locality to preserve overall node representations, which is consistent with the accuracy analysis in Table~\ref{tab: prelim_exp_res}. (3) The proposed EGNN shows the most flattened loss landscape than MLP
and GNNs, which implied that EGNN can preserve overall node representations better than other
model architectures.

\subsection{Why ENN performs so bad}


Below we experimentally analyze why ENN performs so bad on the graph dataset.
The key idea of ENN is to fine-tune the model a few steps to make it prepare for editing.
Specifically, it is explicitly designed to make every sample closer to the decision boundary. 
In this way, the wrongly predicted samples are easier to be perturbed across the boundary.
However, we found that this extra fine-tuning process significantly hurts the model performance.
As shown in Table \ref{tab: enn_detail}, we report the test accuracy for the baseline (i.e., before editing), the accuracy after fine-tuned by ENN, and the accuracy after editing.
We summarize that there was a significantly accuracy drop after fine-tuned by ENN, which significantly compromises the model's performance to prepare it for editing

\begin{table}
\centering
\caption{The test accuracy (\%) for detailed ENN performance analysis}
\label{tab: enn_detail}
\begin{tabular}{cccccc} 
\toprule
Model                      & Method                                                        & Cora  & A-computers & A-photo & Coauthor-CS  \\ 
\hline
\multirow{3}{*}{GCN}       & Baseline                                                      & 89.4  & 87.88       & 93.77   & 94.37        \\ 
\cline{2-6}
                           & \begin{tabular}[c]{@{}c@{}}After ENN \\fine-tune\end{tabular} & 32.0  & ~52.97      & 9.70    & 1.92         \\ 
\cline{2-6}
                           & After Edits                                                   & 37.16 & 15.51       & 16.71   & 4.94         \\ 
\hline
\multirow{3}{*}{GraphSAGE} & Baseline                                                      & 86.6  & 82.83       & 94.30   & 95.17        \\ 
\cline{2-6}
                           & \begin{tabular}[c]{@{}c@{}}After ENN\\fine-tune\end{tabular}  & 32.00 & 7.00        & 4.60    & 13.06        \\ 
\cline{2-6}
                           & After Edits                                                   & 33.16 & 16.89       & 15.06   & 13.71        \\
\bottomrule
\end{tabular}
\end{table}

\section{Theoretical Analysis on Why Model editing may Cry}
\label{app: analsis}
To deeply understand why model editing may cry in GNNs, we provide a pilot theoretical analysis on one-layer GCN and one-layer MLP for binary node classification task.
Specifically, we consider the model prediction be defined as $f^{GCN}_{\Theta}(\bf{X})=\sigma(\Tilde{\bf{A}}\bf{X}\Theta)$ and $f^{MLP}_{\Theta}(\bf{X})=\sigma(\bf{X}\Theta)$, where $\sigma(\cdot)$ is sigmoid activation function, $\bf{X}\in\mathbb{R}^{n\times d}$, and $\Theta\in\mathbb{R}^{d\times 1}$. Then we have the following informal statement:
\begin{theorem}[Informal]\label{theo:informal}
    For well-trained one-layer GCN $f^{GCN}_{\Theta_1}$ and one-layer MLP $f^{MLP}_{\Theta_2}$ for binary node classification task, suppose GCN has sharp locality loss landscape than MLP, model editing (parameters fine-tuning) incurs higher KL divergence locality loss for $f^{GCN}_{\Theta_1}$ than $f^{MLP}_{\Theta_2}$.
\end{theorem}
\paragraph{Remark:} Theorem~\ref{theo:informal} represents that model editing in GNNs leads to higher prediction differences than that of MLPs. Note that such analysis is only based on one-layer model with a binary node classification task, we leave the analysis for more complicated cases (e.g., multi-layer models, and multi-class classification) for future work.

We only consider well-trained one-layer GCN and MLP for binary classification task, defined as $f^{GCN}_{\Theta_1}(\bf{X})=\sigma(\Tilde{\bf{A}}\bf{X}\Theta_1)$ and $f^{MLP}_{\Theta_2}(\bf{X})=\sigma(\bf{X}\Theta_2)$, where $\sigma(\cdot)$ is sigmoid activation function, $\bf{X}\in\mathbb{R}^{n\times d}$, and $\Theta_1, \Theta_2\in\mathbb{R}^{d\times 1}$. Define the training nodes index set as $\mathcal{V}_{train}$ and the misclassified node index as $j$, where $j\notin \mathcal{V}_{train}$. We use $\hat{y}_i$ to represent the model prediction of node $v_i$ for GCN or MLP models, and add superscript
to indicate a specific model.
We use cross-entropy loss for misclassified node $v_j$ in model editing and use gradient descent to update model parameters, i.e.,
\begin{equation}\label{eq:editpara}
    \Theta' = \Theta - \alpha \frac{\partial \mathcal{L}_{CE}(y_j, \hat{y}_j)}{\partial \Theta},
\end{equation}
where $\alpha$ is step size, cross-entropy loss is $\mathcal{L}_{CE}(y_i, \hat{y}_i)=-y_i\log \hat{y}_i - (1-y_i)\log(1-\hat{y}_i)$. We define $\hat{y}'_i$ to represent the model prediction of node $v_i$ after model editing. We adopt the KL divergence between after and before model editing to measure the locality of the well-trained model, i.e.,
\begin{equation}
    \mathcal{L}_{KL} = \frac{1}{|\mathcal{V}_{train}|}\sum_{i\in\mathcal{V}_{train}}\mathcal{L}_{KL}(\hat{y}'_i, \hat{y}_i)=\frac{1}{|\mathcal{V}_{train}|}\sum_{i\in\mathcal{V}_{train}}\hat{y}'_i\log\frac{\hat{y}'_i}{\hat{y}_i}+(1-\hat{y}'_i)\log\frac{(1-\hat{y}'_i)}{(1-\hat{y}_i)},
\end{equation}
The main goal is to compare the KL locality $\mathcal{L}_{KL}^{GCN}$ and $\mathcal{L}_{KL}^{GCN}$ for GCN and MLP model resulting from model editing with parameters update. Note that the model parameters update is relatively small, the KL locality can be effectively approximated using one-order Taylor expansion.

Note that $\mathcal{L}_{KL}=0$ if $\Theta'=\Theta$ and model editing only leads to small model parameters perturbations, we can expand $\mathcal{L}_{KL}$ as follows:
\begin{equation}
    \mathcal{L}_{KL} = \frac{1}{|\mathcal{V}_{train}|}\sum_{i\in\mathcal{V}_{train}}\frac{\partial \mathcal{L}_{KL}(\hat{y}'_i, \hat{y}_i)}{\partial  \Theta'}\big\|_{ \Theta'= \Theta}(\Theta'-\Theta) + (\Theta'-\Theta)^{\top}\mathbf{H}\big\|_{ \Theta'= \Theta}(\Theta'-\Theta) + o(\|\Theta'-\Theta\|^2_F)
\end{equation}
where Hessian matrix $\mathbf{H}\big\|_{\Theta'=\Theta}=\frac{\partial^2 \mathcal{L}_{KL}(\hat{y}'_i, \hat{y}_i)}{\partial ( \Theta')^2}\big\|_{ \Theta'= \Theta'}$.
We omit the term $o(\|\Theta'-\Theta\|_F)$ due small model parameter perturbations in the following analysis. Notice that the derivative of sigmoid function is $\frac{\partial \sigma(x)}{x}=\sigma(x)\big(1-\sigma(x)\big)$, the first derivative of KL locality for the individual sample can be given as
\begin{align}
    \frac{\partial \mathcal{L}_{KL}(\hat{y}'_i, \hat{y}_i)}{\partial  \Theta'}\big\|_{ \Theta'= \Theta}&=\frac{\partial \mathcal{L}_{KL}(\hat{y}'_i, \hat{y}_i)}{\partial \hat{y}'_i}\big\|_{\hat{y}'_i=\hat{y}_i}\frac{\partial \hat{y}'_i}{\partial \Theta'}\big\|_{ \Theta'= \Theta} \nonumber\\
    &= \Big(\log\frac{\hat{y}'_i}{\hat{y}_i}-\log\frac{(1-\hat{y}'_i)}{(1-\hat{y}_i)}\Big)\big\|_{\hat{y}'_i=\hat{y}_i}\frac{\partial \hat{y}'_i}{\partial \Theta'}\big\|_{ \Theta'= \Theta}=\mathbf{0}.
\end{align}
Therefore, the main part to analyze locality loss $\mathcal{L}_{KL}$ is Hessian matrix $\mathbf{H}\big\|_{\Theta'=\Theta}$. For simplicity, we first consider MLP model, and the first derivative of KL locality for the individual sample can be given as
\begin{align}
    \frac{\partial \mathcal{L}_{KL}(\hat{y}'_i, \hat{y}_i)}{\partial  \Theta'}=\Big(\log\frac{\hat{y}'_i}{\hat{y}_i}-\log\frac{(1-\hat{y}'_i)}{(1-\hat{y}_i)}\Big)\hat{y}'_i(1-\hat{y}'_i)\mathbf{X}_{i, :}^\top\triangleq g(\hat{y}'_i)\mathbf{X}_{i, :}^\top
\end{align}
It is easy to obtain that
\begin{align}
    \frac{\partial g(\hat{y}'_i)}{\partial \hat{y}'_i}\big\|_{\hat{y}'_i=\hat{y}_i} = (\frac{1}{\hat{y}'_i}+\frac{1}{1-\hat{y}'_i})\hat{y}'_i(1-\hat{y}'_i)+\Big(\log\frac{\hat{y}'_i}{\hat{y}_i}-\log\frac{(1-\hat{y}'_i)}{(1-\hat{y}_i)}\Big)(1-2\hat{y}'_i)\big\|_{\hat{y}'_i=\hat{y}_i} =1
\end{align}
Therefore, we have Hessian matrix
\begin{align}
    \mathbf{H}^{MLP}\big\|_{\Theta'=\Theta}&=\frac{\partial^2 \mathcal{L}^{MLP}_{KL}(\hat{y}'_i, \hat{y}_i)}{\partial ( \Theta')^2}\big\|_{ \Theta'= \Theta'}=\frac{\partial g(\hat{y}'_i)}{\partial \hat{y}'_i} \hat{y}'_i(1-\hat{y}'_i)\mathbf{X}_{i, :}^\top\mathbf{X}_{i, :} \nonumber\\
    &=\hat{y}'_i(1-\hat{y}'_i)\mathbf{X}_{i, :}^\top\mathbf{X}_{i, :}
\end{align}
The  locality of the well-trained MLP model for individual node $v_i$ is approximately given by
\begin{align}
    \mathcal{L}_{KL}(\hat{y}'_i, \hat{y}_i) = (\Theta'-\Theta)^{\top}\mathbf{H}\big\|_{ \Theta'= \Theta}(\Theta'-\Theta) = \hat{y}'_i(1-\hat{y}'_i)\|\mathbf{X}_{i, :}(\Theta'-\Theta)\|^2.
\end{align}
Note that cross-entropy loss for
misclassified node $v_j$ is adopted in model editing and model parameters update via gradient descent, ground-truth $y_j$ is given by $y_j=-u(\hat{y}_j-0.5)$, where $u(\cdot)$ is a step function, and $\frac{\partial \mathcal{L}_{CE}(y_j, \hat{y}_j)}{\partial \hat{y}_j}=-\frac{y_j}{\hat{y}_j}+\frac{1-y_j}{1-\hat{y}_j}=\frac{u(\hat{y}_j-0.5)}{\min\{\hat{y}_j, 1-\hat{y}_j\}}$, the model editing gradient is given by
\begin{align}\label{eq:editgrad}
    \frac{\partial \mathcal{L}_{CE}(y_j, \hat{y}_j)}{\partial \Theta} &= \frac{u(\hat{y}_j-0.5)}{\min\{\hat{y}_j, 1-\hat{y}_j\}}\hat{y}_j(1-\hat{y}_j)\mathbf{X}_{j, :}^\top \nonumber\\
    &=u(\hat{y}_j-0.5)\max\{\hat{y}_j, 1-\hat{y}_j\}\mathbf{X}_{j, :}^\top
\end{align}
The locality of the well-trained MLP model for individual node $v_i$ can be simplified as
\begin{align}
    \mathcal{L}^{MLP}_{KL}(\hat{y}'_i, \hat{y}_i) = \hat{y}'_i(1-\hat{y}'_i)\max\{\hat{y}_j, 1-\hat{y}_j\}\langle\mathbf{X}_{i, :}, \mathbf{X}_{j, :}\rangle
\end{align}
The average locality of the well-trained MLP model for training nodes is
\begin{align}\label{eq:locality_MLP}
    \mathcal{L}^{MLP}_{KL} = \frac{1}{|\mathcal{V}_{train}|}\sum_{i\in\mathcal{V}_{train}}\hat{y}'_i(1-\hat{y}'_i)\max\{\hat{y}_j, 1-\hat{y}_j\}\langle\mathbf{X}_{i, :}, \mathbf{X}_{j, :}\rangle
\end{align}
As for GCN model, the only difference from MLP is node feature aggregation. The average locality of the well-trained GCN model for training nodes can be obtained by replacing $\mathbf{X}$ with $\mathbf{\Tilde{A}}\mathbf{X}$, i.e., 
\begin{align}\label{eq:locality_GCN}
    \mathcal{L}^{GCN}_{KL} = \frac{1}{|\mathcal{V}_{train}|}\sum_{i\in\mathcal{V}_{train}}\hat{y}'_i(1-\hat{y}'_i)\max\{\hat{y}_j, 1-\hat{y}_j\}\langle[\mathbf{\Tilde{A}}\mathbf{X}]_{i, :}, [\mathbf{\Tilde{A}}\mathbf{X}]_{j, :}\rangle
\end{align}

On the other hand, neighborhood aggregation leads node features more similar. According to~\citep[Proposition 1]{oonograph}, suppose graph data has $M$ connected components and $\lambda_1\leq\cdots\leq\lambda_n$ is the eigenvalue of $\mathbf{\Tilde{A}}$ sorted in ascending order, then we have $-1< \lambda_1, \lambda_{n-M}<1$, and $\lambda_{n-M+1}=\cdots=\lambda_{n}=1$. We mainly focus on the largest less-than-one eigenvalue defined as $\lambda\triangleq \max\limits_{k=1, \cdots, n-M}|\lambda_k|<1$. Additionally, define 
subspace $\mathcal{M}\subseteq\mathbb{R}^{n\times d}$ be the linear subspace where all row vectors are equivalent, the over-smoothing issue can be measured using the distance between node feature matrix $\mathbf{X}$ and subspace $\mathcal{M}$ by $d_{\mathcal{M}}(\mathbf{X})\triangleq \inf\{\|\mathbf{X}-\mathbf{Y}\|_F|\mathbf{Y}\in\mathcal{M}\}$. Based on ~\citep[Theorem 2]{oonograph} and $\lambda<1$, we have 
\begin{align}\label{eq:distcomp}
    d_{\mathcal{M}}(\mathbf{\Tilde{A}}\mathbf{X})\leq \lambda d_{\mathcal{M}}(\mathbf{X}) < d_{\mathcal{M}}(\mathbf{X}),
\end{align}
Note that if the raw vector of $\mathbf{Y}$ is the average row vector of $\mathbf{X}$, the distance between node feature matrix $\mathbf{X}$ and subspace $\mathcal{M}$ is given by
\begin{align}
    d_{\mathcal{M}}(\mathbf{X}) &= \sum_{i=1}^{n}\|\mathbf{X}_{i,:} - \frac{1}{n}\sum_{i=1}^{n}\mathbf{X}_{i,:} \|_F = \sum_{i=1}^{n}\| \frac{1}{n} (\mathbf{X}_{i,:} - \sum_{k\neq i}\mathbf{X}_{k,:})  \|_F \nonumber\\
    &=\frac{1}{n^2}\big(\sum_{i=1}^{n}\|\mathbf{X}_{i,:}\|_F - \sum_{i\neq j} \langle \mathbf{X}_{i,:}, \mathbf{X}_{j,:}^\top \rangle\big),
\end{align}
Note that the adjacency matrix is normalized and the scale of the node features matrix is the same, i.e., $\|\mathbf{X}\|_F\approx \|\mathbf{\Tilde{A}}\mathbf{X}\|_F$. Therefore, the distance between node feature matrix $\mathbf{X}$ and subspace $\mathcal{M}$ is inversely related to node feature inner products. 
Based on Eqs (\ref{eq:locality_MLP}), (\ref{eq:locality_GCN}), and (\ref{eq:distcomp}), we have $\mathcal{L}^{MLP}_{KL}< \mathcal{L}^{GCN}_{KL}$, i.e., editable training in one-layer GCN leads to higher prediction differences than that of one-layer MLP.

\end{document}